\documentclass{article}
\usepackage{caption}
\PassOptionsToPackage{numbers, compress}{natbib}

\usepackage{microtype}
\usepackage{graphicx}
\usepackage{makecell}
\usepackage{booktabs} 

\usepackage{hyperref}
\usepackage[table]{xcolor}
\definecolor{aliceblue}{RGB}{176,223,229}
\newcommand{\CC}{\cellcolor{aliceblue}}

\usepackage[ruled, lined, linesnumbered, commentsnumbered, longend]{algorithm2e}

\usepackage[final]{neurips_2024}

\usepackage{amsmath}
\usepackage{amssymb}
\usepackage{mathtools}
\usepackage{amsthm}
\usepackage{subcaption}  
\usepackage{wrapfig}
\theoremstyle{plain}
\newtheorem{theorem}{Theorem}[section]
\newtheorem{proposition}[theorem]{Proposition}

\theoremstyle{definition}

\theoremstyle{remark}

\definecolor{C0}{rgb}{0.121569, 0.466667, 0.705882}
\definecolor{C1}{rgb}{1.000000, 0.498039, 0.054902}
\definecolor{C2}{rgb}{0.172549, 0.627451, 0.172549}
\definecolor{C3}{rgb}{0.839216, 0.152941, 0.156863}
\definecolor{C4}{rgb}{0.580392, 0.403922, 0.741176}
\definecolor{C5}{rgb}{0.549020, 0.337255, 0.294118}
\definecolor{C6}{rgb}{0.890196, 0.466667, 0.760784}
\definecolor{C7}{rgb}{0.498039, 0.498039, 0.498039}
\definecolor{C8}{rgb}{0.737255, 0.741176, 0.133333}
\definecolor{C9}{rgb}{0.090196, 0.745098, 0.811765}
\newcommand\ve[1]{\mathbf{#1}}

\usepackage[textsize=tiny]{todonotes}

\title{MagicDistillation: \textit{Weak-to-Strong} Video Distillation for Large-Scale Few-Step Synthesis}

\author{	
Shitong Shao$^{1,2,*}$ ~~~ Hongwei Yi$^{2,*,\dagger}$ ~~~ Hanzhong Guo$^{2,3}$ ~~~ Tian Ye$^{1,2}$ ~~~ Daquan Zhou$^{4}$ \\ \\
\textbf{Michael Lingelbach}$^{2}$ ~~~ \textbf{Zhiqiang Xu}$^{5}$ ~~~ \textbf{Zeke Xie}$^{1,\ddagger}$ \\ \\
$^1$ HKUST(GZ)  \quad $^2$ Hedra Inc. \quad $^3$ HKU \quad $^4$ Peking University \quad $^5$ MBZUAI \\
}

\begin{document}

\maketitle

\begin{abstract}
Recently, open-source video diffusion models (VDMs), such as WanX, Magic141 and HunyuanVideo, have been scaled to over 10 billion parameters. These large-scale VDMs have demonstrated significant improvements over smaller-scale VDMs across multiple dimensions, including enhanced visual quality and more natural motion dynamics. However, these models face two major limitations: (1) High inference overhead: Large-scale VDMs require approximately 10 minutes to synthesize a 28-step video on a single H100 GPU. (2) Limited in portrait video synthesis: Models like WanX-I2V and HunyuanVideo-I2V often produce unnatural facial expressions and movements in portrait videos. To address these challenges, we propose MagicDistillation, a novel framework designed to reduce inference overhead while ensuring the generalization of VDMs for portrait video synthesis. Specifically, we primarily use sufficiently high-quality talking video to fine-tune Magic141, which is dedicated to portrait video synthesis. We then employ LoRA to effectively and efficiently fine-tune the fake DiT within the step distillation framework known as distribution matching distillation (DMD). Following this, we apply weak-to-strong (W2S) distribution matching and minimize the discrepancy between the fake data distribution and the ground truth distribution, thereby improving the visual fidelity and motion dynamics of the synthesized videos. Experimental results on portrait video synthesis demonstrate the effectiveness of MagicDistillation, as our method surpasses Euler, LCM, and DMD baselines in both FID/FVD metrics and VBench. Moreover, MagicDistillation, requiring only 4 steps, also outperforms WanX-I2V (14B) and HunyuanVideo-I2V (13B) on visualization and VBench. {Our project page is \textcolor{C6}{\href{https://magicdistillation.github.io/MagicDistillation/}{\textcolor{C6}{https://magicdistillation.github.io/MagicDistillation/}}}}.
\end{abstract}

\renewcommand{\thefootnote}{}
\footnotetext{$^{*}$Equal contribution, $^{\dagger}$Project lead, $^\ddagger$Corresponding author.}
\renewcommand{\thefootnote}{\arabic{footnote}}

\section{Introduction}
\label{sec:introduction}
As generative intelligence gradually transitions from image generation to video synthesis~\citep{SD35,SDV15,llama,animatediff,genmo2024mochi}, an increasing number of high-quality large-scale video diffusion models (VDMs) have emerged. Compared to traditional VDMs built on U-Net~\citep{UNet} (\textit{e.g.}, AnimateDiff~\citep{animatediff} and ModelScope-T2V~\citep{modelscope}), these models, leveraging diffusion transformers (DiTs)~\citep{DIT}, have made groundbreaking advances in the quality of synthesized videos. However, even though these VDMs adhere well to physical laws and exhibit strong generalization capabilities for video synthesis across a variety of scenarios, the primary challenges associated with those VDMs is the significant inference overhead. For instance, on an advanced GPU like H100, synthesizing a 28-step, 129-frame (\textit{i.e.}, $\sim\!5$s) video with flash attention3~\citep{flashattn3} using HunyuanVideo (13B)~\citep{kong2024hunyuanvideo}, Magic141~\citep{yi2025magic} or WanX (14B)~\citep{wan2025} takes approximately 10 minutes, rendering them completely unsuited for real-time applications.

\begin{figure*}[t]
        \centering
        \includegraphics[width=\linewidth,trim={0 0 0cm 0},clip]{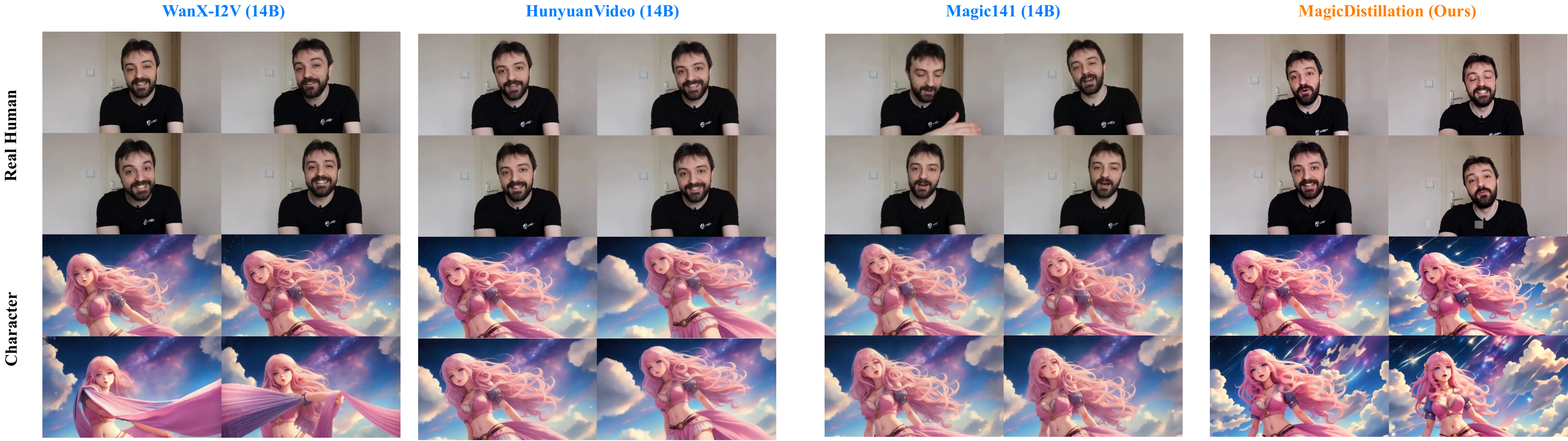} 
        \vspace{-15pt}
        \caption{Comparative visualization of synthesized videos for both real humans and characters using WanX-I2V (14B), HunyuanVideo-I2V (14B), Magic141, and MagicDistillation. Notably, WanX-I2V, HunyuanVideo-I2V, and Magic141 results are generated using 50 sampling steps, while MagicDistillation achieves comparable results with only 4 sampling steps.}
        \label{fig:prior_comparison}
        \vspace{-18pt}
\end{figure*}

Apart from the substantial inference GPU latency that these VDMs entail, they are still woefully insufficient in synthesizing portrait videos~\citep{zakharov2020fast,cui2024hallo3}, which encompass both real humans and characters, and remains notably inadequate. As illustrated in Fig.~\ref{fig:prior_comparison}, WanX-I2V and HunyuanVideo-I2V demonstrate evident shortcomings in motion dynamics when it comes to portrait video synthesis. Moreover, WanX-I2V exhibits issues with inconsistency in character identity. It should be noted that, historically, there have been several algorithms specifically designed to address portrait video synthesis, such as EMO~\citep{EMO}, Hallo-2~\citep{cui2024hallo2}, and Hallo-3~\citep{cui2024hallo3}. However, these algorithms have compromised their ability to synthesize general videos.

\begin{figure}[t]
    \centering
    \vspace{-5pt}
    \begin{subfigure}[t]{0.40\textwidth}
        \centering
        \includegraphics[width=\linewidth,trim={0 0 0 0},clip]{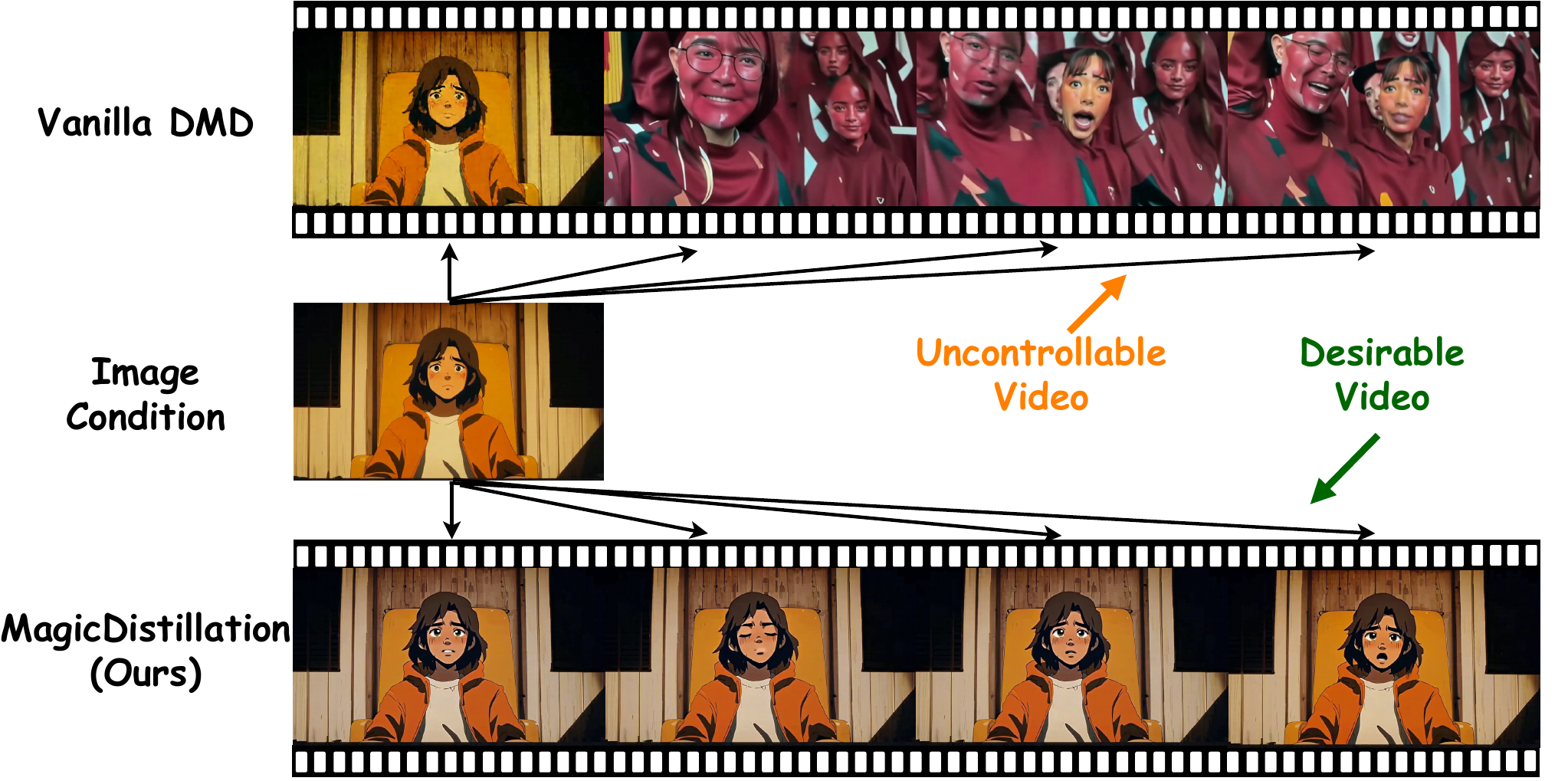} 
        \vspace{-10pt}
        \label{fig:pre_visualization}
    \end{subfigure}
    \hfill
    \begin{subfigure}[t]{0.56\textwidth} 
        \centering
        \includegraphics[width=\linewidth,trim={0 0 1.15cm 0},clip]{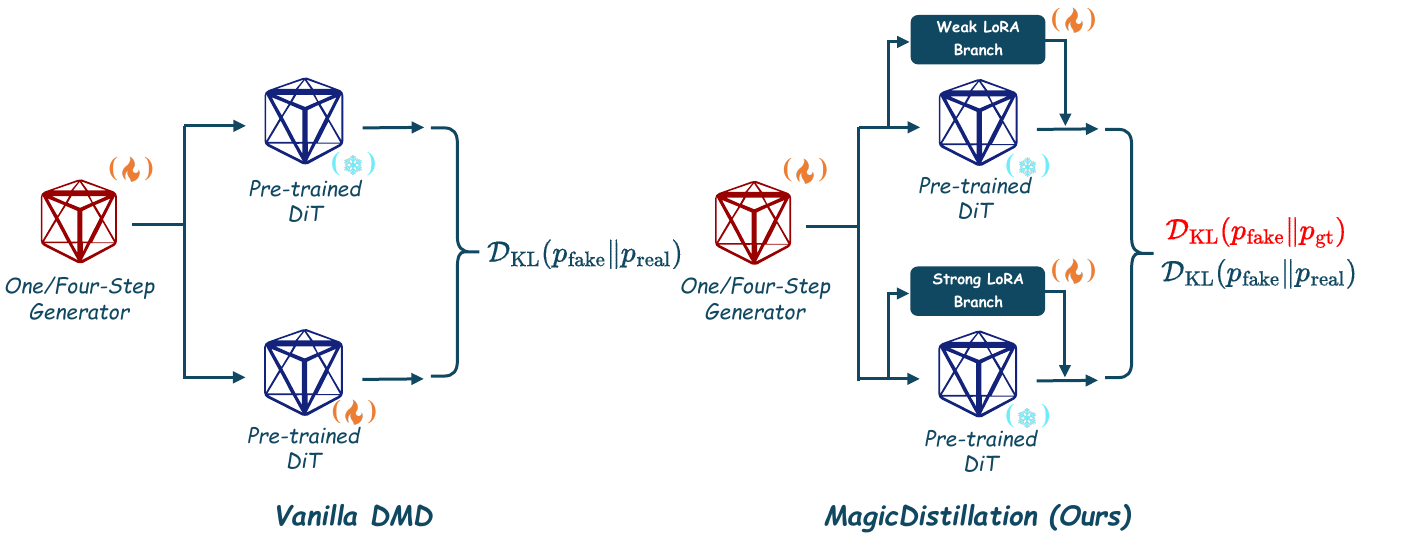} 
        \vspace{-20pt}
        \label{fig:w2s_dmd_easy_framework}
    \end{subfigure}
    \vspace{-5pt}
    \caption{
    Left: Vanilla DMD suffers from training collapse (4-step models with 900 iterations). 
    Right: Our framework integrates LoRA for \textit{weak-to-strong} distribution matching and $\mathcal{D}_\textrm{KL}$ constraints.
    } 
    \label{fig:pre_combined_results}
    \vspace{-20pt}
\end{figure}

To address the significant inference delays associated with large-scale VDMs and their inability to generate high-quality videos of realistic humans, full-body figures, and stylized anime characters, this paper investigates the utilization of Magic141~\citep{yi2025magic}, a model trained jointly on general video and talking video datasets, as a teacher model for step distillation~\citep{mcm_accelerate,I2V_turbo}. Step distillation significantly enhances the efficiency of video synthesis by drastically reducing the number of function evaluations (NFEs). For practical implementation considerations, particularly regarding GPU memory usage and training efficiency (details provided in Appendix~\ref{apd:choice_dmd}), we adopt distribution matching distillation (DMD) to carry out step distillation. To further enhance the efficiency of the distillation process, as presented in Fig.~\ref{fig:pre_combined_results} (Right), we apply low-rank adaptation (LoRA) to the real model (\textit{i.e.}, estimating the real data distribution) in order to construct the fake model (\textit{i.e.}, estimating the fake data distribution).

Unfortunately, we observed that employing LoRA to implement vanilla DMD presents two major issues: \textcolor{C3}{\textit{1)}} unstable training, which is highly prone to overfitting and leads to uncontrollable synthesized video outputs (see Fig.~\ref{fig:pre_combined_results} (Left)), as well as \textcolor{C3}{\textit{2)}} insufficient visual quality, as evidenced by metrics such as VBench~\citep{vbench,vbench++}, FID~\citep{fid} and FVD~\citep{unterthiner2019fvd} (see Sec.~\ref{sec:experiment}). We hypothesize that the root cause of these issues lies in the fact that, as training progresses, the real model gradually fails to adequately accommodate the inputs synthesized by the generator, leading to inaccuracies in the KL divergence approximation. To address this issue, as illustrated in Fig.~\ref{fig:pre_combined_results} (Right), we propose \textbf{MagicDistillation}, a method that subtly adjusts the parameters of the real model toward those of the fake model. This adjustment enhances the real model's ability to accommodate the few-step generator's outputs, thereby improving training stability. We also introduce a KL divergence approximation between the fake data distribution and the ground truth distribution (\textit{i.e.}, $\mathcal{D}_\textrm{KL}(p_\textrm{fake}\Vert p_\textrm{gt})$ in Fig.~\ref{fig:pre_combined_results} (Right), $\mathcal{D}_\textrm{KL}$ stands for the KL divergence) to further enhance both subject generalization and the visual fidelity of synthesized videos. This serves as an optional regularization term, designed to guide the data distribution of the few-step generator closer to the ground truth data distribution, which can enhance the fidelity while also improving the motion dynamics.


We validate our proposed approach using benchmark datasets, our own collected widescreen video dataset and the general I2V-VBench~\citep{vbench++}. For the first one, we conduct evaluations using both FID and FVD metrics on benchmark datasets such as VFHQ~\citep{vfhq}, HTDF~\citep{htdf}, and Celeb-V~\citep{celebv}, showcasing that MagicDistillation achieves a significant performance leap over vanilla DMD/DMD2 and LCM. For the second one, we construct a new VBench, a benchmark specifically designed for evaluating portrait video synthesis, using widescreen portrait images (see Appendix~\ref{apd:benchmark}). The experimental results reveal that our 4-step model outperforms other methods in 6 out of 7 metrics, surpassing the best performance achieved by vanilla DMD2. For the last one, by leveraging higher-quality talking video and general video datasets, we trained a more effective 4-step model. We compared this model against WanX-I2V (14B), HunyuanVideo-I2V (13B), and Magic141. The results of our customized VBench and general I2V-VBench~\citep{vbench++} demonstrate that MagicDistillation outperforms these VDMs, even when they use 50 sampling steps.

\begin{figure*}
    \vspace{-5pt}
    \centering
    \includegraphics[width=1.0\linewidth]{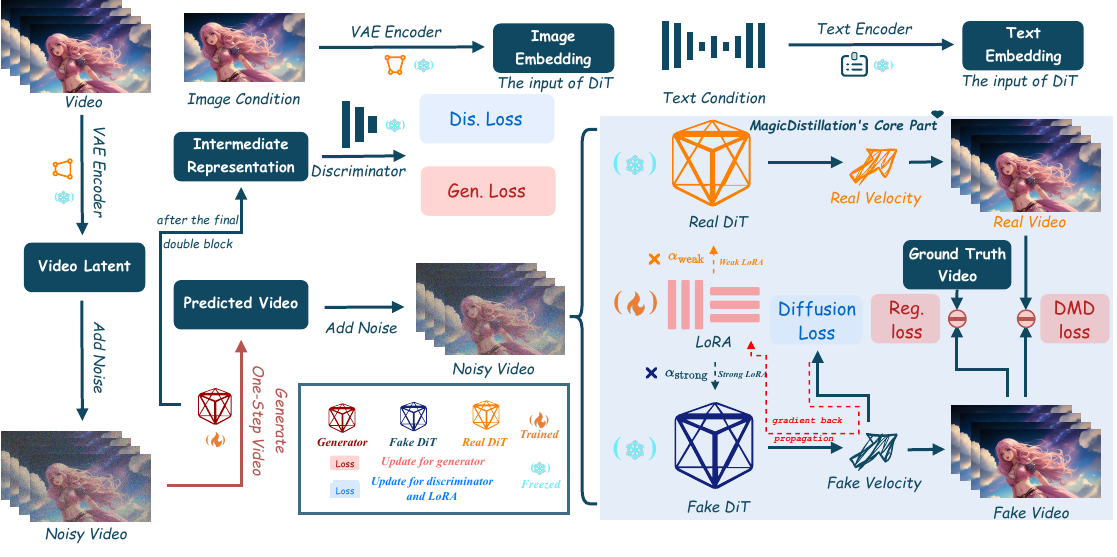}
    \vspace{-20pt}
    \caption{Illustration of MagicDistillation. ``Reg.'', ``Dis.'', and ``Gen.'' stand for ``Regularization'', ``Discriminator'', and ``Generator'', respectively. MagicDistillation primarily leverages LoRA to facilitate the training of a large-scale VDM. The weight factors $\alpha_\textrm{strong}$ and $\alpha_\textrm{weak}$ are employed to achieve the W2S distribution matching. Furthermore, the regularization loss, which incorporates the ground truth video, helps to alleviate the overfitting problem encountered with the DMD Loss.}
    \label{fig:w2s_dmd_framework}
    \vspace{-16pt}
\end{figure*}

\section{Related Work}
\label{sec:related_work}

Here, we briefly review the scheduler as well as the architecture of the large-scale open-source VDMs, the step distillation algorithm and portrait video synthesis.

\noindent{\bf Large-Scale Open-Source VDMs.} WanX~\citep{wan2025}, Step-Video~\citep{stepfunvideo}, HunyuanVideo~\citep{kong2024hunyuanvideo}, and Magic141~\citep{yi2025magic} are widely acknowledged as some of the most advanced and efficient open-source VDMs. Their schedulers employ the flow-matching paradigm~\citep{flow_matching} to model both the forward and reverse processes. To be specific, flow matching defines a mapping between (Gaussian) noises $\ve{x}_1$ from a noise distribution $p_1(\ve{x}_1)$ to samples $\ve{x}_0$ from a real data distribution $p_0(\ve{x}_0)$ in terms of an ordinary differential equation (ODE): $d\ve{x}_t = \ve{v}_\Theta(\ve{x}_t,t) dt$, where $\ve{x}_t$ and $\ve{v}_\Theta(\cdot,\cdot)$ denote the sample at timestep $t$ on the flow and the velocity estimation model, respectively. Flow matching optimizes $\ve{v}_\Theta(\cdot,\cdot)$ by solving a simple least squares regularization problem:
\begin{equation}
    \begin{split}
    & \mathop{\arg\min}_{\Theta} \int_0^1 \mathbb{E}\left[\alpha_t\Vert \partial p_t(\ve{x}_t)/\partial t - \ve{v}_\Theta(\ve{x}_t,t)\Vert_2^2\right]dt, \\
    \end{split}
    \label{eq:hunyuan2}
\end{equation}
where $\alpha_t$ is the weight factor, which takes the form of logit-normal by default~\citep{karras2022elucidating}. After training, sampling is done directly using the Euler algorithm. Moreover, the backbone of these VDMs are all constructed based on DiT rather than UNet, effectively increasing their scaling potential and performance ceiling. Notably, VAEs utilized by these VDMs share an identical compression ratio for video input. They downsample from $T\!\times\!H\!\times\!W$ to $\lfloor\!\frac{T}{4}\!\rfloor\!\times\!\lfloor\frac{H}{8}\!\rfloor\!\times\!\lfloor\!\frac{W}{8}\!\rfloor$, where $T$, $H$, and $W$ denote the temporal length, height, and width of the video, respectively.


\noindent{\bf Step Distillation.} Step distillation enables accelerated sampling by reducing the NFEs during inference of the diffusion model. At present, the prevailing paradigms in this domain revolve around two main approaches: ``few steps aligned to many steps'' and ``distribution matching''. Among these, LCM stands out as the primary framework for aligning few-step models with multi-step processes. A majority of the open-source few-step VDMs are constructed using LCM training, including MCM~\citep{mcm_accelerate}, the T2V-turbo series~\citep{t2vturbov2}, and FastVideo~\citep{fastvideo}. The paradigm ``distribution matching'' has demonstrated superior performance over LCM in image synthesis, as substantiated by methods such as DMD~\citep{dmd}, DMD2~\citep{dmd2}, and SiD-LSG~\citep{sid_lsg}. However, its application within video synthesis remains relatively uncommon~\citep{yin2024slow}.

\noindent{\bf Portrait Video Synthesis.} The core objective of portrait video synthesis is to synthesize realistic talking-head videos by ensuring that the synthesized output aligns seamlessly with the reference image, audio, or text inputs. Previous algorithms predominantly rely on UNet as the backbone alongside diffusion models for portrait video synthesis. While these approaches achieve competitive performance, such as EMO~\citep{EMO}, Sadtalker~\citep{sadtalker} and Hallo~\citep{cui2024hallo2}, they face significant challenges in scaling to longer videos and maintaining high video fidelity. This essentially stems from the lack of scalability in UNet, coupled with the insufficient length of videos used during the training process.


\section{Method}
\label{sec:method}
As illustrated in Fig.~\ref{fig:w2s_dmd_framework}, we present MagicDistillation for accelerated sampling of large-scale portrait video synthesis. MagicDistillation incorporates \textit{weak-to-strong} (\textit{W2S}) distribution matching along with ground truth supervision. This technique aims to prevent the synthesis of uncontrollable videos and to improve visual quality in the few-step video synthesis scenario. Moreover, we develop a tailored discriminator specifically for our pre-trained VDM Magic141 and put forward a more efficient algorithm designed for the 1-step generator distillation.

\subsection{Preliminary}

In this part, we provide a brief overview of the distribution matching distillation (DMD). Let $G_\phi$ denote the few-step generator parameterized by $\phi$, and let two auxiliary models, $\ve{v}_\theta^\textrm{fake}$ and $\ve{v}_\Theta^\textrm{real}$, be initialized from a pre-trained diffusion model. The central concept of DMD is to minimize the reverse KL divergence across randomly sampled timesteps $t$ between the real data distribution $p_\textrm{real}$ obtained from $\ve{v}_\Theta^\textrm{real}$ and the fake data distribution $p_\textrm{fake}$ obtained from $\ve{v}_\theta^\textrm{fake}$:\begin{equation}
\fontsize{9pt}{11pt}\selectfont
    \begin{split}
        &\nabla_\phi \mathcal{L}_\textrm{DMD} \triangleq \mathbb{E}_{t} \nabla_\phi \mathcal{D}_\textrm{KL}\left(p_\textrm{fake}\Vert p_\textrm{real}\right) \approx - \mathbb{E}_t \int_\epsilon\Big([\ve{x}_t  -\sigma_t\ve{v}_\Theta^\textrm{real}(\ve{x}_t,t) -  \ve{x}_t+\sigma_t\ve{v}_\theta^\textrm{fake}(\ve{x}_t,t)]\frac{\partial G_\phi(\epsilon)}{\partial \phi}d\epsilon\Big), \\
    \end{split}
    \label{eq:dmd_loss}
\end{equation}where $\ve{x}_t = (1-\sigma_t)G_\phi(\epsilon)+\sigma_t \epsilon$. Here, $\epsilon$ and $\sigma_t$ stand for the Gaussian noise and the noise schedule, respectively. Note that this form is the result of converting DMD from DDPM~\citep{ddpm_begin} (\textit{w.r.t}, predict noise) to flow matching (\textit{w.r.t}, predict velocity). During training, to dynamically align $p_\textrm{fake}$ with the output distribution of $G_\phi$ in real time, we minimize the following loss:
\begin{equation}
    \begin{split}
    \mathcal{L}_\textrm{diffusion} & = \mathbb{E}_{t\sim \mathcal{U}[0,1],\epsilon^\prime\sim \mathcal{N}(0,\mathbf{I})}\Big[\alpha_t\big\Vert \epsilon^\prime - G_\phi(\epsilon) -\ve{v}^\textrm{fake}_\theta(\sigma_t\epsilon^\prime+(1-\sigma_t)G_\phi(\epsilon),t)\big\Vert_2^2\Big], \\
    \end{split}
    \label{eq:diffusion_loss}
\end{equation}where $\epsilon^\prime$ and $\epsilon$ are Gaussian noises and independent of each other. By alternately optimizing Eq.~\ref{eq:dmd_loss} and Eq.~\ref{eq:diffusion_loss}, we obtain a few-step generator capable of synthesizing high-quality videos. Moreover, the key improvement of DMD2 over DMD lies in the inclusion of a discriminator $D_\xi$, which is designed to further enhance video quality. This paradigm adopts an alternating optimization strategy, where $D_\xi$ and $G_\phi$ are iteratively optimized during the DMD training phase, which can be formulated as
\begin{equation}
    \begin{split}
        & \mathcal{L}_\textrm{dis} = \mathbb{E}_{\epsilon,\ve{x}^\textrm{gt}}\max \left(0, 1\!+\!D_\xi(G_\phi(\epsilon))\right) + \max \left(0, 1\!-\!D_\xi(\ve{x}^\textrm{gt})\right),\quad \mathcal{L}_\textrm{gen} = - \mathbb{E}_\epsilon D_\xi(G_\phi(\epsilon)),\\
    \end{split}
    \label{eq:adv_loss}
\end{equation}where $\ve{x}^\textrm{gt}$ refers to the ground truth video, and $\mathcal{L}_\textrm{dis}$ and $\mathcal{L}_\textrm{gen}$ are used to update the discriminator $D_\xi$ and the few-step generator $G_\phi$, respectively.

\subsection{\textit{Weak-to-Strong} Distribution Matching}
To introduce the \textit{weak-to-strong} (\textit{W2S}) distribution matching, we should further rewrite and understand the DMD optimization problem (\textit{i.e.}, minimize Eq.~\ref{eq:dmd_loss}, subject to Eq.~\ref{eq:diffusion_loss}):
\begin{equation}
\fontsize{9pt}{11pt}\selectfont
    \begin{split}
        & \mathop{\arg\min}_{\phi}\ \mathbb{E}_{t,(\epsilon,\epsilon^\prime)\sim \mathcal{N}(0,\mathbf{I})}\Vert\epsilon^\prime - G_\phi(\epsilon) - \ve{v}_\Theta^\textrm{real}(\ve{x}_t,t)\Vert_2^2.
    \end{split}
    \label{eq:the_int}
\end{equation}\begin{table*}[!t]
\centering
\vskip -0.01in
\small
\caption{\small {I2V quantitative results comparison using Magic141 (trained on both portrait and general video datasets) on our customized VBench. All videos are 540$\times$960$\times$129 and are saved at 24fps. Each sample follow VBench~\citep{vbench} to synthesize 5 videos to avoid errors.}}
\scalebox{0.66}{
\begin{tabular}{lccccccccc}
\Xhline{3\arrayrulewidth}\\[-2ex]
{\textbf{Method}} & {\textbf{\makecell{i2v\\subject} (↑)}} & {\textbf{\makecell{subject\\consistency} (↑)}} & {\textbf{\makecell{motion\\smoothness} (↑)}} & {\textbf{\makecell{dynamic\\degree} (↑)}} & {\textbf{\makecell{aesthetic\\quality} (↑)}} & {\textbf{\makecell{imaging\\quality} (↑)}} & {\textbf{\makecell{temporal\\flickering} (↑)}} &  {\textbf{average (↑)}}\\
\Xhline{3\arrayrulewidth}
      \\[-2ex]
        \multicolumn{9}{l}{\large\textit{\textbf{28-step}}}\\\Xhline{3\arrayrulewidth}
{Euler (baseline)} & 0.9578 & 0.8933 & 0.9947 & 0.7000 & 0.5278 & 0.6106 & 0.9906 &  0.8107 \\
\\[-2ex]
\multicolumn{9}{l}{\large\textit{\textbf{4-step}}}\\\Xhline{3\arrayrulewidth}
{Euler (baseline)}  & 0.9738 & 0.8773 & \CC 0.9967 & 0.0448 & 0.4574 & 0.3902 & 0.9971 &  0.6768 \\
LCM & 0.8809 & 0.9313 & 0.9897 & 0.0551 & 0.4729 & 0.5250 & 0.9855 & 0.6915 \\
DMD2 & 0.9613 & 0.9111 & 0.9932 & 0.7586 & 0.4643 & 0.5725 & 0.9832 & 0.8063 \\
{MagicDistillation w/o reg. loss} & \CC 0.9763 & \CC 0.9533 & 0.9954 & \CC 0.7683 & 0.4899 & 0.5708 & \CC 0.9993 & \CC 0.8219\\
{MagicDistillation w/ reg. loss} & 0.9735 & 0.9478 & 0.9955 & 0.0379 & \CC 0.5732 & \CC 0.7103 & 0.9939 & 0.7474 \\
\\[-2ex]
\multicolumn{9}{l}{\large\textit{\textbf{1-step}}}\\\Xhline{3\arrayrulewidth}
{Euler (baseline)} & 0.8729 & 0.8177 & \CC 0.9964 & 0.0000 & 0.3906 & 0.2441 & \CC 0.9968 & 0.6169\\
LCM & 0.9662 & 0.9236 & 0.9963 & 0.0310 & 0.4465 & 0.4125 & 0.9959 & 0.6817 \\
DMD2 & \CC 0.9742 & 0.9590 & 0.9963 & 0.0054 & 0.5090 & 0.5922 & 0.9964 &  0.7189 \\
{MagicDistillation w/o reg. loss} & 0.9620 & \CC 0.9656 & 0.9961 & \CC 0.0523 & \CC 0.5512 & \CC 0.6909 & 0.9953 & \CC 0.7448 \\
\Xhline{3\arrayrulewidth}\\[-2ex]
\end{tabular}
}
\label{tab:comparison}
\vskip -0.07in
\end{table*}\begin{figure*}[t]
    \vspace{-5pt}
    \centering
    \includegraphics[width=1.0\linewidth]{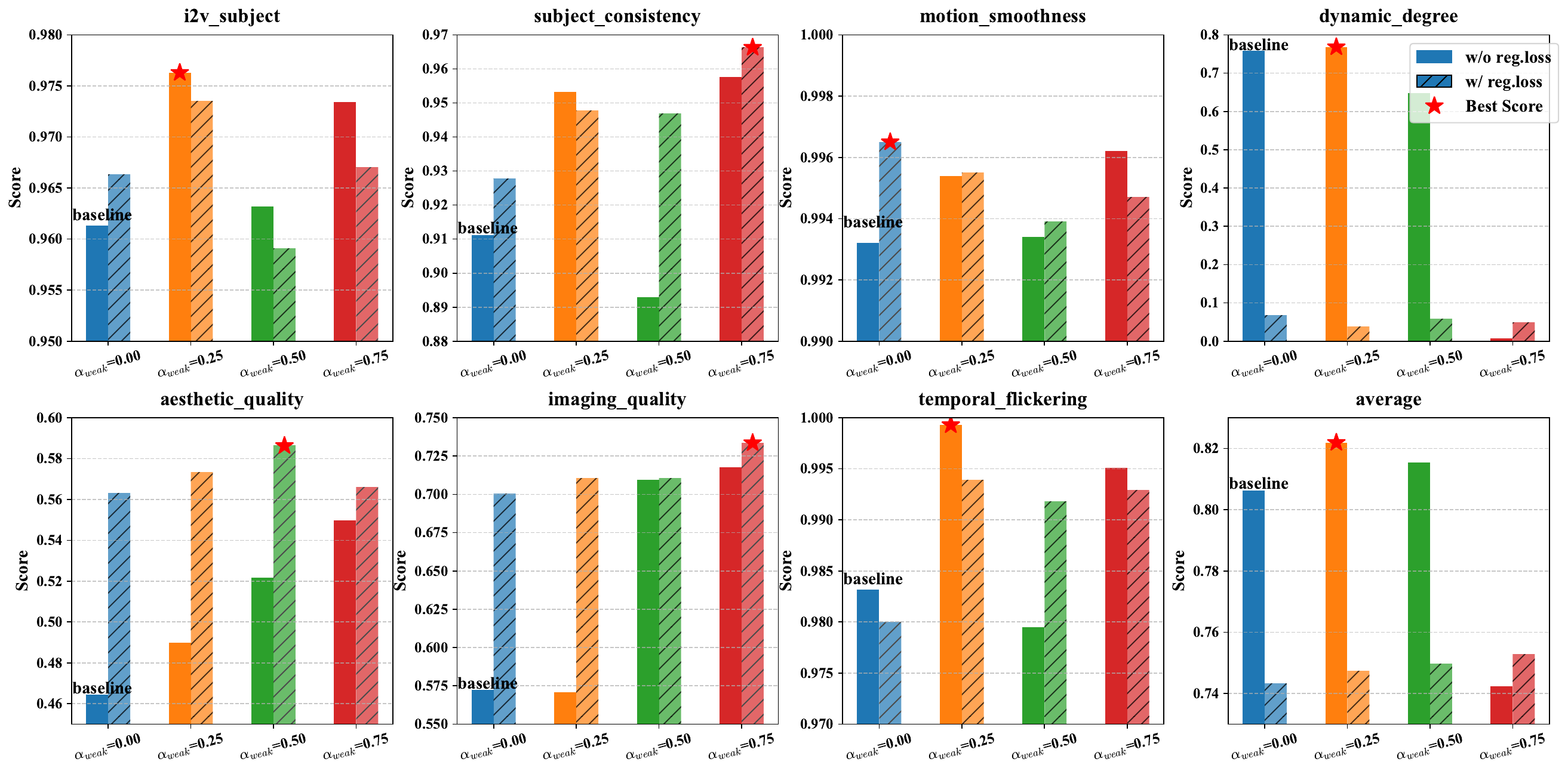}
    \vspace{-18pt}
    \caption{Ablation studies of $\alpha_\textrm{weak}$ and $\mathcal{L}_\textrm{reg}$ using 4-step models on our customized VBench. MagicDistillation reduces to the vanilla DMD2 when $\alpha_\textrm{weak}$=0. From the average metrics presented in the lower right corner, it is evident that MagicDistillation achieves its optimal performance when  $\alpha_\textrm{weak}$=0.25. Furthermore, when the visual quality of the ground truth video is high but the motion dynamic is insufficient, the regularization loss (\textit{i.e.}, $\mathcal{L}_\textrm{reg}$) represents a trade-off between motion dynamics and visual quality.}
    \label{fig:ablation_studies}
    \vspace{-15pt}
\end{figure*}\begin{wrapfigure}{r}{5.3cm}
    \vspace{-5pt}
    \centering
    \includegraphics[width=1.0\linewidth]{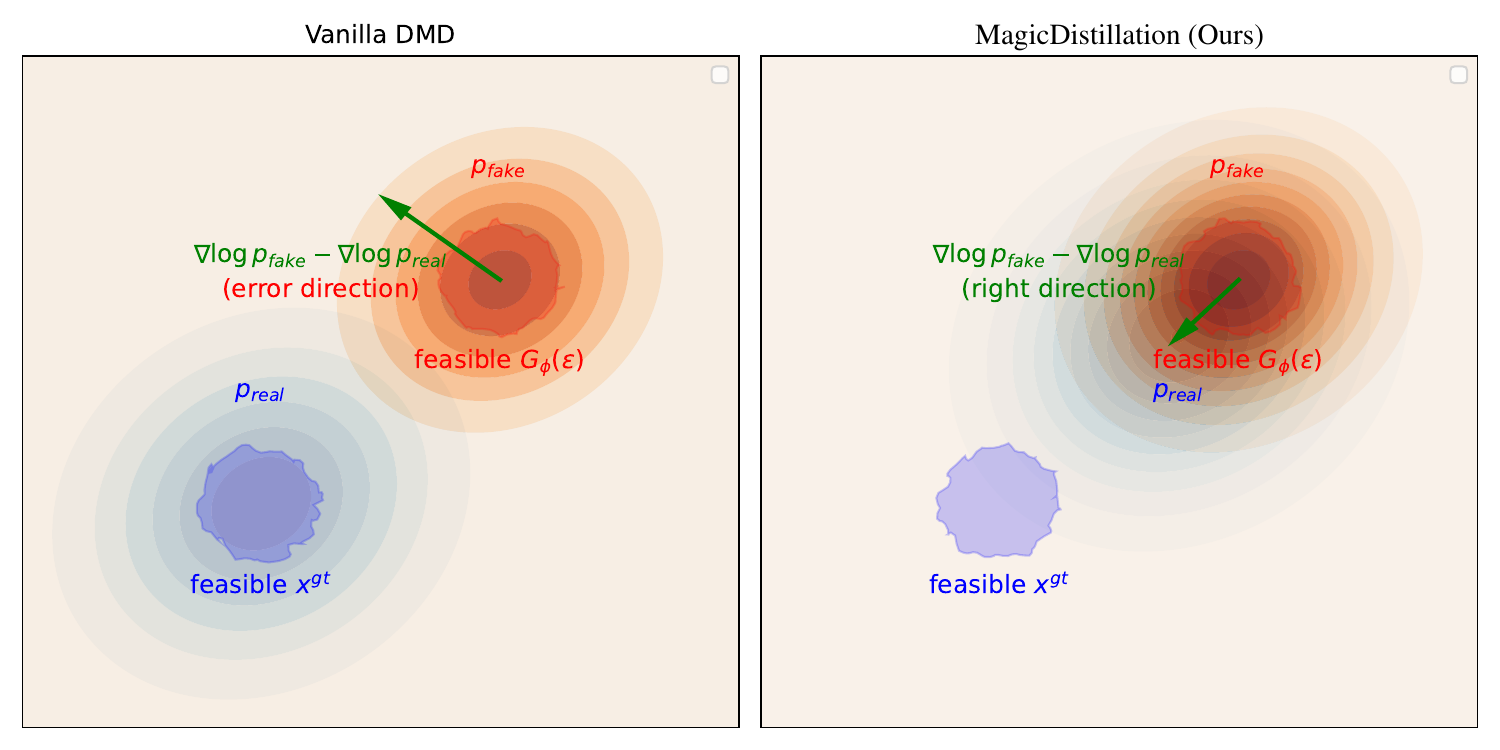}
    \vspace{-15pt}
    \caption{\small {Vanilla DMD vs. MagicDistillation. Vanilla DMD encounters a significant challenge that $p_\textrm{real}$ has no overlap with the feasible region of the sample synthesized by the few-step generator $G_\phi$, which leads to an inaccurate estimation of the gradient difference $\nabla \log p_\textrm{fake} - \nabla \log p_\textrm{real}$. In contrast, MagicDistillation mitigates this issue by subtly shifting $p_\textrm{real}$ toward $p_\textrm{fake}$. This technique adjustment ensures a more substantial overlap between the two distributions, thereby enhancing the accuracy of the gradient estimation process.}}
    \label{fig:w2s_dmd_motivation}
    \vspace{-12pt}
\end{wrapfigure}Suppose $\Theta^* = \mathop{\arg\min}_{\Theta}\mathbb{E}_{t,\ve{x}^\textrm{gt}}[\Vert \epsilon - \ve{x}^\textrm{gt} - \ve{v}_\Theta^\textrm{real}((1-\sigma_t)\ve{x}_\textrm{gt}+\sigma_t \epsilon,t)\Vert]$ is the optimal pre-trained diffusion model. In practical scenarios, it is inherently infeasible to optimize $\Theta$ to its optimal solution $\Theta^*$. Thus, Eq.~\ref{eq:the_int} can only theoretically make $G_\phi(\epsilon)$ as close as possible to the data distribution $p_\textrm{real}$ estimated by $\ve{v}^\textrm{real}_\Theta$. Furthermore, we can obtain $\ve{v}_\theta^\textrm{fake}(\ve{x}_t,t)=\epsilon^\prime - G_\phi(\epsilon)$ by minimizing Eq.~\ref{eq:diffusion_loss} and $\ve{v}_\Theta^\textrm{real}(\ve{x}_t,t)=\epsilon^\prime - G_\phi(\epsilon)$ by optimizing Eq.~\ref{eq:the_int}, leading to the conclusion that $\Theta\equiv\theta$. Note that at the initial stage of DMD training, $\theta$ is equal to $\Theta$. Consequently, during the optimization process, $\theta$ initially diverges from $\Theta$ and subsequently converges back closer to $\Theta$. The above analysis indicates that the distance between $\theta$ and $\Theta$ remains relatively small throughout the training process. This observation makes it a compelling choice to incorporate LoRA for optimizing $\theta$, as it can effectively reduce computational overheads.

\begin{table*}[!t]
\centering
\vskip -0.01in
\small
\caption{\small {TI2V quantitative results comparison between MagicDistillation and other populer large-scale VDMs on our customized VBench. In particular, the VDM referenced for 28-step sampling using the Euler (baseline) in Table~\ref{tab:comparison} corresponds to Magic141 as presented in this table.}}
\scalebox{0.66}{
\begin{tabular}{lccccccccc}
\Xhline{3\arrayrulewidth}\\[-2ex]
{\textbf{Method}} & {\textbf{\makecell{i2v\\subject} (↑)}} & {\textbf{\makecell{subject\\consistency} (↑)}} & {\textbf{\makecell{motion\\smoothness} (↑)}} & {\textbf{\makecell{dynamic\\degree} (↑)}} & {\textbf{\makecell{aesthetic\\quality} (↑)}} & {\textbf{\makecell{imaging\\quality} (↑)}} & {\textbf{\makecell{temporal\\flickering} (↑)}} &  {\textbf{average (↑)}}\\
\Xhline{3\arrayrulewidth}
      \\[-2ex]
        \multicolumn{9}{l}{\large\textit{\textbf{50-step}}}\\\Xhline{3\arrayrulewidth}
{WanX-I2V (14B)} & 0.9800 & 0.9649 & 0.9910 & 0.5625 & 0.5980 & 0.7031 & 0.9858  &  0.8264 \\
{HunyuanVideo-I2V (13B)} & \CC 0.9834 & \CC 0.9654 & \CC 0.9959 & 0.1639 & \CC 0.5992 & 0.6990 & \CC 0.9940  &  0.7715 \\
{Magic141 (13B)} & 0.9677 & 0.9607 & 0.9936 & 0.7704 & 0.5892 & 0.6906 & 0.9895 &  0.8516 \\
\\[-2ex]
\multicolumn{9}{l}{\large\textit{\textbf{4-step}}}\\\Xhline{3\arrayrulewidth}
{MagicDistillation}  & 0.9659 & 0.9374 & 0.9881 & \CC 0.9935 & 0.5782 & \CC 0.7122 & 0.9727 &  \CC 0.8782 \\
\Xhline{3\arrayrulewidth}\\[-2ex]
\end{tabular}
}
\label{tab:comparison_2}
\vskip -0.14in
\end{table*}
\begin{figure*}[!t]
    \vspace{-5pt}
    \centering
    \includegraphics[width=1.0\linewidth]{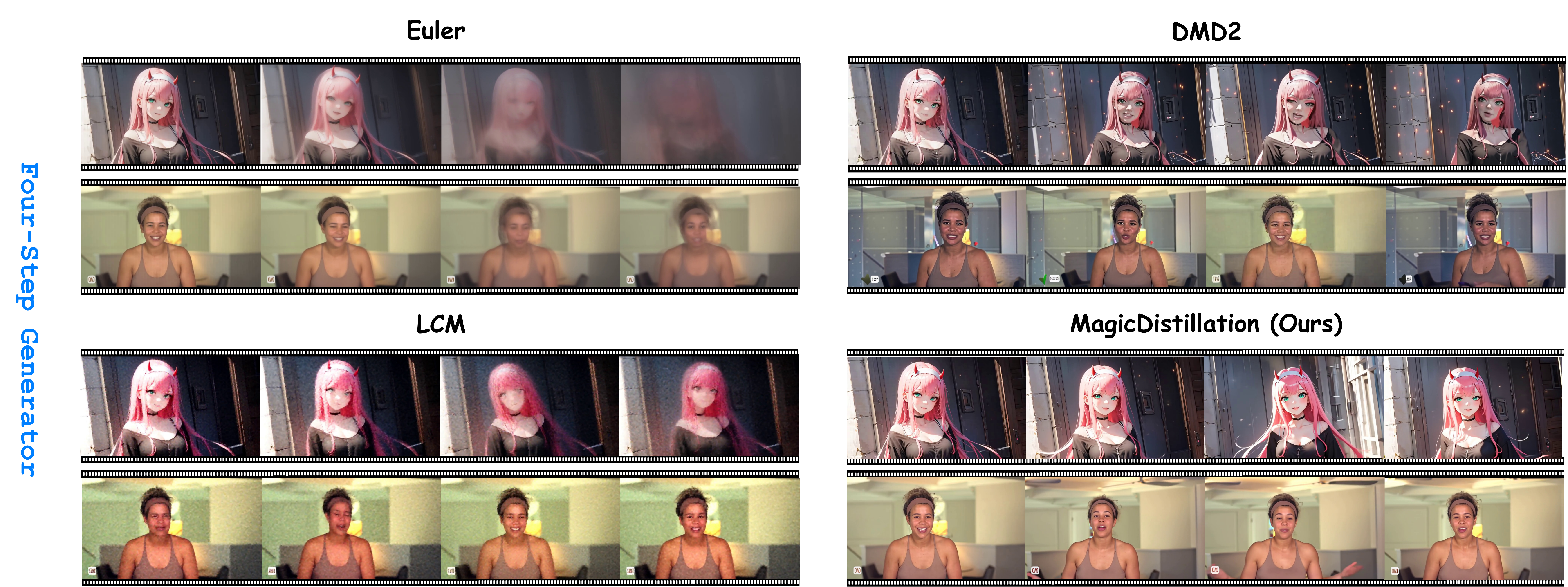}
    \vspace{-15pt}
    \caption{Visualization of various accelerated sampling methods under the 4-step scenario. Additional visualizations for the 1-step scenario and other details are provided in Appendix~\ref{apd:add_visualization}.}
    \label{fig:post_visualization}
    \vspace{-14pt}
\end{figure*}

Another advantage of incorporating LoRA is that \textit{W2S} distribution matching can be achieved by adjusting the weight factor. As illustrated in Fig.~\ref{fig:w2s_dmd_motivation}, this method addresses the issue where fails to overlap $p_\textrm{real}$ with the feasible region of samples synthesized by the few-step generator $G_\phi$. To be specific, we observe that Magic141 exhibits a large discrepancy between $p_\textrm{real}$ and $p_\textrm{fake}$ during DMD training for portrait video synthesis tasks, as illustrated on the left side of Fig.~\ref{fig:w2s_dmd_motivation}. This mismatch leads to an imprecise estimation of $\nabla \log p_\textrm{fake} - \nabla \log p_\textrm{real}$, ultimately synthesizing videos that lack controllability, as shown in Fig.~\ref{fig:pre_combined_results} (Left). Intuitively, this issue typically emerges when the difference between $\Theta$ and $\theta$ surpasses the acceptable tolerance range. A practical solution to address this problem is to slightly adjust $p_\textrm{real}$ to better align it with $p_\textrm{fake}$. This adjustment can be effectively achieved by adjusting the LoRA weight factors $\alpha_\textrm{weak}$ (used for $\ve{v}_\Theta^\textrm{real}$) and $\alpha_\textrm{strong}$ (used for $\ve{v}_\theta^\textrm{fake}$):
\begin{equation}
    \begin{split}
        & \ve{v}^\textrm{real}_\Theta(\ve{x}_t,t) = \alpha_\textrm{weak} \zeta(\ve{x}_t,t) + \ve{v}_\Theta^\textrm{pre-train}(\ve{x}_t,t),\\
        & \ve{v}_\theta^\textrm{fake}(\ve{x}_t,t) = \alpha_\textrm{strong}\zeta(\ve{x}_t,t) + \ve{v}_\Theta^\textrm{pre-train}(\ve{x}_t,t),\\
    \end{split}
\end{equation}where $\ve{v}_\Theta^\textrm{pre-train}$ denotes the pre-trained diffusion model, and $\zeta$ stands for the low-rank branch. Obviously, in \textit{W2S} distribution matching Eq.~\ref{eq:diffusion_loss}, is employed to optimize LoRA. When $\alpha_\textrm{weak}=0$, the \textit{W2S} distribution matching simplifies to the standard distribution matching, which relies on LoRA to optimize $\ve{v}_\theta^\textrm{fake}$. We fix $\alpha_\textrm{strong}$ to be 1 by default in our experiments and obtain the empirically optimal solution by appropriately tuning $\alpha_\textrm{weak}$.

\begin{proposition}
\label{the:w2s_dmd_work}
(the proof in Appendix~\ref{apd:the_1}) The optimization objective of \textit{W2S} distribution matching remains identical to that of the standard distribution matching, with the low-rank branch $\zeta(\ve{x}_t,t)$ serving as the intermediate teacher to facilitate better optimization.
\vspace{-3pt}
\end{proposition}

We further leverage Proposition~\ref{the:w2s_dmd_work} to demonstrate that \textit{W2S} distribution matching not only ensures training stability but also achieves the same optimization objective as the standard distribution matching. Notably, as LoRA serves as an intermediate teacher, \textit{W2S} distribution matching also enhances the performance of step distillation. Our empirical results (see Sec.~\ref{sec:experiment}) show that \textit{W2S} distribution matching surpasses vanilla DMD in terms of motion dynamics, semantic faithfulness and visual quality. By utilizing higher-quality portrait and general video datasets for step distillation, MagicDistillation, with a mere 4-step sampling process (\textit{i.e.}, NFE$=$4), is capable of outperforming WanX-I2V, HunyuanVideo-I2V, and Magic141 in portrait video synthesis, even under their 50-step sampling scenarios (\textit{i.e.}, NFE$=$100).

\subsection{Ground Truth Supervision}
Another essential component of MagicDistillation is ground truth supervision, which is designed to enhance the stability of distribution matching while improving the visual quality of synthesized videos. The few-step generator is expected to match distributions not only between $p_\textrm{real}$ and $p_\textrm{fake}$ but also between $p_\textrm{gt}$ (\textit{i.e.}, $p(\ve{x}^\textrm{gt})$) and $p_\textrm{fake}$. This matching serves as a corrective measure against the challenges of unstable training process and the degradation of visual quality caused by the inherent biases present in $\ve{v}_\Theta^\textrm{pre-train}$. In our implementation. we minimize the following KL divergence $\mathcal{L}_\textrm{reg}$:\begin{equation}
\fontsize{7pt}{11pt}\selectfont
    \begin{split}
        &\nabla_\phi \mathcal{L}_\textrm{reg} \triangleq \mathbb{E}_{t} \nabla_\phi \mathcal{D}_\textrm{KL}\left(p_\textrm{fake}\Vert p_\textrm{gt}\right) = \nabla_\phi\frac{1}{2}\mathbb{E}[\Vert \ve{x}^\textrm{gt} - \ve{x}_t + \sigma_t \ve{v}_\theta^\textrm{fake}(\ve{x}_t,t)\Vert_2^2] \\
& \approx - \mathbb{E}_t \int_\epsilon\Big([\ve{x}^\textrm{gt} -  \ve{x}_t +\sigma_t\ve{v}_\theta^\textrm{fake}(\ve{x}_t,t)]\frac{\partial[\ve{x}_t - \sigma_t\ve{v}_\theta^\textrm{fake}(\ve{x}_t,t)]}{\partial G_\phi(\epsilon)}\frac{\partial G_\phi(\epsilon)}{\partial \phi}d\epsilon\Big)  \\
        & = - \mathbb{E}_t \int_\epsilon\Big([\ve{x}^\textrm{gt} -  \ve{x}_t +\sigma_t\ve{v}_\theta^\textrm{fake}(\ve{x}_t,t)]\frac{\partial G_\phi(\epsilon)}{\partial \phi}d\epsilon\Big),\\
        &\quad s.t.\quad \mathcal{L}_\textrm{diffusion} \leq \eta, \\
    \end{split}
    \label{eq:reg_loss}
\end{equation}where $\eta$ stands for a very small amount. As demonstrated in Sec.~\ref{sec:experiment}, we demonstrate that $\mathcal{L}_\textrm{reg}$ is effective in pulling together the distribution of the synthesized video obtained from few-step generator and the distribution of the ground truth video.

\vspace{-5pt}
\subsection{Detailed Implementation}

\noindent{\bf Tailored Discriminator.} Adversarial training is a useful technique for enhancing the performance of step distillation. However, traditional discriminator architectures have predominantly relied on convolution-based designs. It is noteworthy that the backbone architecture of Magic141 consists of multi-model (MM) DiT blocks, which serves as the foundation for the double blocks in Magic141, alongside vanilla DiT blocks, which forms the single blocks in the corresponding counterpart. To better align with the structure of Magic141, we incorporate a DiT-based discriminator. Specifically, we use the output from the final double block as the input for the discriminator and employ the initial layer of the single blocks, capitalizing on pre-trained weights while ensuring that no parameters are shared, as the foundation for the discriminator's architecture.

\noindent{\bf 1-Step Generator Training.} Using synthesized noise-data pairs for distillation~\citep{iclr22_rect} can significantly enhance the performance of the 1-step generator. However, synthesizing noise-data pairs using Magic141 with 50 sampling steps is computationally expensive and time-intensive. To mitigate this, we employ the 4-step generator obtained from MagicDistillation to synthesize noise-data pairs. Since the computational cost of 4-step sampling is relatively low, this process can be seamlessly integrated into the MagicDistillation training pipeline, eliminating the need for pre-synthesizing noise-data pairs. Initially, we set the distillation loss weight to 1 and the loss weight of $\mathcal{L}_\textrm{DMD}$ to 0.25. During the training of the 1-step generator, we progressively adjust the distillation~\citep{iclr22_rect} loss weight to 0.25 while increasing the loss weight of $\mathcal{L}_\textrm{DMD}$ to 1, which can ensure a balanced optimization process.

\noindent{\bf LoRA in Fake DiT and Real DiT.} We find that when utilizing bfloat16 precision within the deepspeed framework~\citep{deepspeed}, the F-norm of the gradient during LoRA fine-tuning is significantly smaller compared to that observed during full-parameter fine-tuning. This discrepancy results in the vanishing gradient issue. Given the constraints imposed by using ZeRO3 in deepspeed, we address this problem by multiplying the loss associated with LoRA by an additional weight factor of 1e4, enabling stable and effective training under normal conditions.

\vspace{-5pt}
\section{Experiment}
\label{sec:experiment}
All experiments involving MagicDistillation, vanilla DMD/DMD2, and LCM for the 4-step generator, with the exception of those detailed in Table~\ref{tab:comparison_2}, Table~\ref{tab:general_comparison} and Table~\ref{tab:comparison_apd}, were carried out using 8 NVIDIA H100 GPUs. In addition to this, MagicDistillation in Table~\ref{tab:comparison_2}, Table~\ref{tab:general_comparison} and Table~\ref{tab:comparison_apd} uses 40 NVIDIA H100 GPUs. For the 1-step generator, experiments with MagicDistillation and vanilla DMD/DMD2 were performed using 16 NVIDIA H100 GPUs. For all datasets except those used for MagicDistillation in Table~\ref{tab:comparison_2}, our training data comprises high-quality widescreen videos sourced from YouTube, carefully curated to match the resolution of 540$\times$960. Additionally, we incorporate a selection of anime data to further enhance the generalization capability of MagicDistillation. For the experimental results showcased in Table~\ref{tab:comparison_2}, we incorporate a selected portion of general data. This data is carefully curated from Koala-36M~\citep{koala_dataset} and Intern4k~\citep{intern4k_dataset}, with stringent conditions to ensure both superior aesthetic quality and dynamic motion attributes.

Our evaluation framework incorporates FID and FVD metrics assessed on the VFHQ~\citep{vfhq}, HDTF~\citep{htdf}, and Celeb-V~\citep{celebv} datasets, the general I2V-VBench~\citep{vbench++}, alongside our customized VBench~\citep{vbench,vbench++}, which includes seven dimensional metrics such as dynamic degree and image quality. Comprehensive details regarding the training hyperparameter configurations and the construction of the benchmarks are provided in Appendix~\ref{apd:hyperparameter} and~\ref{apd:benchmark}, respectively.

We employ Magic141~\citep{yi2025magic} as the teacher model to perform step distillation. Magic141, a 13B open-source large-scale VDM, is developed through the co-training of portrait video and general video datasets. This model serves as both our I2V-teacher and TI2V-teacher. For I2V synthesis, we simplify the prompt by setting it to null during model training and performance evaluation. It is worth noting that the teacher model referenced in Tables~\ref{tab:comparison},~\ref{tab:hdtf},~\ref{tab:vfhq} and~\ref{tab:celebv}, along with the few-step generators such as DMD/DMD2, LCM, and MagicDistillation, are exclusively I2V-generators. Besides, the models referenced in the remaining tables are all TI2V-generators. Further experimental results pertaining to the TI2V-generator are comprehensively detailed in Appendix~\ref{apd:ti2v_generator}.

\begin{table*}[!t]
\centering
\vskip -0.01in
\small
\caption{\small {TI2V quantitative results comparison between MagicDistillation and other popular large-scale VDMs on general I2V-VBench. 
Experimental results for the other comparison methods were directly extracted from the official VBench website. Note that the results for WanX-I2V and HunyuanVideo-I2V are not presented here, as the official website does not provide data for these VDMs.}}
\scalebox{0.77}{
\begin{tabular}{lcccccccc}
\Xhline{3\arrayrulewidth}\\[-2ex]
{\textbf{Method}} & {\textbf{\makecell{i2v\\subject} (↑)}} & {\textbf{\makecell{subject\\consistency} (↑)}} & {\textbf{\makecell{motion\\smoothness} (↑)}} & {\textbf{\makecell{dynamic\\degree} (↑)}} & {\textbf{\makecell{aesthetic\\quality} (↑)}} & {\textbf{\makecell{imaging\\quality} (↑)}} &  {\textbf{average (↑)}}\\
\Xhline{3\arrayrulewidth}
      \\[-2ex]
        \multicolumn{8}{l}{\large\textit{\textbf{100-NFE}}}\\\Xhline{3\arrayrulewidth}
CogVideoX-I2V-SAT~\citep{yang2024cogvideox} & \CC 0.9767 & 0.9547 & 0.9835 & 0.3651 & 0.5976 & 0.6764  & 0.7590 \\
I2Vgen-XL~\citep{2023i2vgenxl} & 0.9752 & 0.9636 & 0.9831 & 0.2496 & 0.6533 & 0.6985  & 0.7538 \\
SEINE-512x320~\citep{chen2023seine} & 0.9657 & 0.9420 & 0.9668 & 0.3431 & 0.5842 & 0.7097  & 0.7519 \\
VideoCrafter-I2V~\citep{chen2024videocrafter2} & 0.9117 & \CC 0.9786 & 0.9800 & 0.2260 & 0.6078 & \CC 0.7168  & 0.7368 \\
SVD-XT-1.0~\citep{blattmann2023stable} & 0.9752 & 0.9552 & 0.9809 & 0.5236 & 0.6015 & 0.6980  & 0.7890 \\
Step-Video-TI2V~\citep{stepfunvideo} & 0.9550 & 0.9602 & \CC 0.9924 & 0.4878 & \CC 0.6299 & 0.7044  & 0.7882 \\
\\[-2ex]
\multicolumn{8}{l}{\large\textit{\textbf{4-NFE}}}\\\Xhline{3\arrayrulewidth}
MagicDistillation  & 0.9512 & 0.9006 & 0.9766 & \CC 0.9840 & 0.5955 & 0.6949  &  \CC 0.8505 \\
\Xhline{3\arrayrulewidth}\\[-2ex]
\end{tabular}
}
\label{tab:general_comparison}
\vskip -0.28in
\end{table*}
\begin{table}[!t]
\centering
\small

\begin{minipage}[t]{0.32\linewidth}
\centering
\renewcommand{\arraystretch}{0.9}
\caption{\footnotesize I2V comparison on VFHQ. Videos are in the shape of 540$\times$960$\times$129.}
\label{tab:vfhq}
\scalebox{0.53}{
\begin{tabular}{lccc}
\toprule
{\textbf{Method}} & {\textbf{NFE (↓)}} & {\textbf{FID (↓)}} & {\textbf{FVD (↓)}} \\
\midrule
Euler (baseline) & 28 & 31.16 & 170.07 \\
Euler (baseline) & 4 & 39.06 & 325.08 \\
LCM & 4 & 76.63 & 598.99 \\
DMD & 4 & 38.41 & 285.95 \\
DMD2 & 4 & 47.49 & 249.22 \\
MagicDistillation w/o reg. loss& 4 & 32.37 & \CC 163.40 \\
MagicDistillation w/ reg. loss& 4 & \CC 31.10 & 239.28 \\
\bottomrule
\end{tabular}}
\end{minipage}
\hfill
\begin{minipage}[t]{0.32\linewidth}
\centering
\renewcommand{\arraystretch}{0.9}
\caption{\footnotesize I2V comparison on HDTF. Videos are in the shape of 540$\times$960$\times$129.}
\label{tab:hdtf}
\scalebox{0.53}{
\begin{tabular}{lccc}
\toprule
{\textbf{Method}} & {\textbf{NFE (↓)}} & {\textbf{FID (↓)}} & {\textbf{FVD (↓)}} \\
\midrule
Euler (baseline) & 28 & 37.75 & 456.39 \\
Euler (baseline) & 4 & 66.79 & 1051.29 \\
LCM & 4 & 79.46 & 1600.59 \\
DMD & 4 & 58.85 & 364.85 \\
DMD2 & 4 & 31.93 & 246.78 \\
MagicDistillation w/o reg. loss& 4 & 24.90 & \CC 233.80 \\
MagicDistillation w/ reg. loss& 4 & \CC 21.92 & 240.02 \\
\bottomrule
\end{tabular}}
\end{minipage}
\hfill
\begin{minipage}[t]{0.32\linewidth}
\centering
\renewcommand{\arraystretch}{0.9}
\caption{\footnotesize I2V comparison on Celeb-V. Videos are in the shape of 540$\times$960$\times$129.}
\label{tab:celebv}
\scalebox{0.53}{
\begin{tabular}{lccc}
\toprule
{\textbf{Method}} & {\textbf{NFE (↓)}} & {\textbf{FID (↓)}} & {\textbf{FVD (↓)}} \\
\midrule
Euler (baseline) & 28 & 72.38 & 698.52 \\
Euler (baseline) & 4 & 110.17 & 1042.60 \\
LCM & 4 & 107.50 & 744.96 \\
DMD & 4 & 72.69 & 661.75 \\
DMD2 & 4 & 104.84 & 682.64 \\
MagicDistillation w/o reg. loss& 4 & 60.85 & \CC 542.72 \\
MagicDistillation w/ reg. loss& 4 & \CC 58.71 & 695.65 \\
\bottomrule
\end{tabular}}
\end{minipage}
\vspace{-12pt}
\end{table}
\subsection{Main Result} 
\label{sec:main_result}
Our comparison experiments were conducted on our customized VBench, as well as on VFHQ, HDTF, and Celeb-V datasets. For VBench, we present the experimental results in Table~\ref{tab:comparison}. Here, the 28-step generator represents the pre-trained Magic141 (\textit{i.e.}, the standard teacher model) specifically adapted for the portrait video synthesis task. Furthermore, the results of Euler (baseline) across all step scenarios are sampled using the pre-trained Magic141. For MagicDistillation, ``w/o reg. loss'' and ``w/ reg. loss'' denote not using ground truth supervision and using ground truth supervision, respectively. In the 4-step scenario, as shown in Table~\ref{tab:comparison}, it can be observed that MagicDistillation (w/o reg. loss) surpasses DMD2 across all metrics except imaging quality, with the average score exceeding DMD2 by a margin of 1.56\%. Furthermore, MagicDistillation (w/o reg. loss) demonstrates a substantial improvement, achieving an increase of over 10\% in the average score compared to both LCM and Euler. In particular, MagicDistillation (w/o reg. loss) even surpasses the 28-step generator (\textit{i.e.}, the teacher model) across all metrics except for imaging quality and aesthetic quality. We attribute this phenomenon to the ability of distribution matching to partially correct the intrinsic bias present in Magic141. Moreover, ground truth supervision can further enhance the visual quality of the synthesized video. Specifically, in Table~\ref{tab:comparison}, MagicDistillation (w/ reg. loss) achieves improvements in imaging quality and aesthetic quality compared to MagicDistillation (w/o reg. loss). Even while the motion dynamics of MagicDistillation (w/ reg. loss) are noticeably lower compared to MagicDistillation (w/o reg. loss), this outcome is primarily attributed to the training data consisting solely of talking videos, which inherently exhibit low motion dynamics. However, when the training data is supplemented with a mix of general videos featuring significantly more pronounced object motion, the motion degree of MagicDistillation, as presented in Table~\ref{tab:comparison_2}, reaches an impressive 0.9935. This value is significantly higher than other compared VDMs. In the 1-step scenario, MagicDistillation surpasses DMD2 across all metrics except for i2v subject, subject consistency, and temporal flickering, achieving a 2.59\% higher average score overall compared to DMD2.

We further compare MagicDistillation, trained with higher-quality general videos, against WanX-I2V, HunyuanVideo-I2V, and Magic141, with the results detailed in Table~\ref{tab:comparison_2} and Table~\ref{tab:general_comparison}. Remarkably, MagicDistillation achieves an outstanding score of 0.8782 on our customized VBench, outperforming all competing methods. Besides, MagicDistillation demonstrates exceptional performance on the general I2V-VBench, as substantiated by the results showcased in Table~\ref{tab:general_comparison}. Its motion degree achieves an impressive 0.9840, significantly surpassing both CogVideoX-I2V-SAT~\citep{cogvideox} and Step-Video-TI2V~\citep{stepfunvideo}. Furthermore, MagicDistillation attains a remarkable final average score of 0.8505, comfortably outperforming the other comparison VDMs, all of which fail to exceed a score of 0.8. The visualized results on {\textcolor{C6}{\href{https://magicdistillation.github.io/MagicDistillation/}{\textcolor{C6}{our project page}}}} further underscore this achievement. Based on these findings, we conclude that even the most renowned open-source VDMs still exhibit notable limitations in portrait video synthesis.

We report the FID and FVD evaluation results on the VFHQ, HDTF, and Celeb-V datasets in Tables~\ref{tab:vfhq},~\ref{tab:hdtf}, and~\ref{tab:celebv}, respectively. As demonstrated by these experimental results, MagicDistillation (w/o reg. loss) outperforms Euler, LCM, and DMD2 in the 4-step scenario across both FID and FVD metrics. Notably, across all datasets and metrics, MagicDistillation (w/o reg. loss) surpasses the 28-step teacher model, with the sole exception of the FID metric on the VFHQ dataset, where it slightly underperforms the 28-step teacher model. Additionally, incorporating ground-truth supervision results in a further reduction in FID but leads to an increase in FVD, which we attribute to the negative impact of ground-truth supervision on motion dynamics when the training data’s motion dynamics are insufficient.
\vspace{-2pt}
\subsection{Ablation Study}
\label{sec:ablation_study}

In this subsection, we perform ablation studies on the hyperparameter $\alpha_\textrm{weak}$ and the loss $\mathcal{L}_\textrm{reg}$. We present the results of the ablation experiments on $\alpha_\textrm{weak}$ and $\mathcal{L}_\textrm{reg}$ in Fig.~\ref{fig:ablation_studies}. When $\alpha_\textrm{weak}$ is set to 0.25, it achieves the best performance in the i2v subject and dynamic degree metrics. Furthermore, as $\alpha_\textrm{weak}$ gradually increases, both aesthetic quality and imaging quality improve monotonically, indicating that increasing $\alpha_\textrm{weak}$ enhances the visual quality of the synthesized videos. Conversely, when $\alpha_\textrm{weak}$ increases from 0.5 to 0.75, the dynamic degree drops sharply, suggesting that an excessively high $\alpha_\textrm{weak}$ negatively impacts the motion dynamics of the synthesized videos. Overall, the average score reaches its optimum when $\alpha_\textrm{weak}=0.25$. Thus, we adopt a default configuration of $\alpha_\textrm{weak} = 0.25$ and $\alpha_\textrm{strong} = 1$. Furthermore, the observation that ground truth supervision substantially enhances both aesthetic quality and imaging quality while adversely affecting the dynamic degree. Based on MagicDistillation’s performance in Tables~\ref{tab:comparison} and~\ref{tab:comparison_2}, we believe that the detrimental effects of $\mathcal{L}_\textrm{reg}$ can be avoided as long as the training data is mixed with general data exhibiting high motion dynamics.
\vspace{-2pt}
\subsection{Visualization}
We provide a visual comparison of Euler, LCM, DMD2, and MagicDistillation in Figs.~\ref{fig:post_visualization} and~\ref{fig:post_visualization_2}. More results can be found in {\textcolor{C6}{\href{https://magicdistillation.github.io/MagicDistillation/}{\textcolor{C6}{project page}}}}. As illustrated in Figs.~\ref{fig:post_visualization} and~\ref{fig:post_visualization_2}, both LCM and Euler sampling produce blurry results in the 1/4-step scenarios, making it difficult to synthesize high-quality talking videos. In contrast, DMD2 synthesizes relatively clearer videos in the 1/4-step scenarios but still suffers from uncontrollable synthesis issues. For example, in the 4-step scenario, the animated portrait videos synthesized by DMD2 exhibit significant changes in the subject's facial features. In comparison, MagicDistillation successfully synthesizes high-fidelity videos in both the 1-step and 4-step scenarios, achieving superior visual quality, subject consistency, and motion dynamics.
\vspace{-2pt}
\section{Conclusion}
\label{sec:conclusion}

In this paper, we introduce a more stable, efficient, and effective step distillation algorithm, termed MagicDistillation, designed for accelerated sampling of large-scale VDM. Our research centers on addressing the substantial inference overhead posed by large-scale VDMs while simultaneously enhancing the performance of few-step generators in the realm of portrait video synthesis. We leveraging the 13B Magic141 model alongside LoRA to facilitate \textit{weak-to-strong} distribution matching. This approach allows for a more reliable estimation of $\mathcal{D}_\textrm{KL}(p_\textrm{fake}\Vert p_\textrm{real})$. Furthermore, we propose incorporating ground truth supervision to achieve superior visual quality. Extensive experimental results demonstrate that MagicDistillation surpasses DMD, DMD2, LCM, the 28/50-step Magic141, the 50-step WanX-I2V and the 50-step HunyuanVideo-I2V in nearly all metrics.

\nocite{langley00}

\clearpage
\paragraph{Ethics Statement.} We present MagicDistillation, a method designed for realizing more efficient and better step distillation on large scale VDMs. The dataset utilized by MagicDistillation is sourced from animated video clips and licensed public portrait videos available online. This dataset undergoes meticulous filtering through both manual review and large language model (LLM)-based processes to ensure that all reference images adhere to ethical standards. Furthermore, we place a strong emphasis on the responsible and ethical deployment of this technology, striving to maximize its societal benefits while diligently addressing and mitigating any potential risks.
\bibliography{icml2025}
\bibliographystyle{icml2025}

\newpage
\appendix
\onecolumn

\section{Benchmarks}
\label{apd:benchmark}

\paragraph{FID and FVD.} In portrait video synthesis, the primary metrics for evaluating video quality are FID~\citep{fid} and FVD~\citep{unterthiner2019fvd}. Following EMO~\citep{EMO} and Hallo~\citep{cui2024hallo2}, we randomly select 100 clips from each of the VFHQ~\citep{vfhq}, HDTF~\citep{htdf}, and Celeb-V~\citep{celebv} datasets. Each clip consists of 129 frames, with the resolution and frame rate standardized to 540$\times$960 and 24 FPS, respectively. We then compute the FID and FVD scores by comparing the synthesized videos with the original videos.

\paragraph{Our Customized VBench.} Evaluating videos solely based on FVD and FID has clear limitations in terms of metric coverage. However, most of the existing video benchmarks, such as Chronomagic-Bench~\citep{yuan2024chronomagic} and T2V-CompBench~\citep{sun2024t2v}, are primarily designed for T2V generation, with only a limited number tailored for I2V synthesis. To address this gap, we adopt the I2V-VBench construction framework from VBench++~\citep{vbench++}. Specifically, we begin by collecting 58 high-quality widescreen images, and use InternVL-26B to generate prompts for these images, and then synthesize five videos per sample (resulting in a total of 290 videos) to minimize testing errors. The synthesized videos are subsequently evaluated using VBench. Note that the key evaluation metrics include: i2v subject, subject consistency, motion smoothness, dynamic degree, aesthetic quality, imaging quality, and temporal flickering.

\paragraph{General I2V-VBench.} The original VBench~\citep{vbench} was designed solely to evaluate the generation performance of T2V-generators and lacked the capability to assess the performance of I2V-generators. Thus, I2V-VBench~\citep{vbench++} is an extension built upon the vanilla VBench framework. This enhancement incorporates over 1118 additional text-image pairs, enabling a comprehensive evaluation of I2V-generators across multiple dimensions, including visual quality, motion dynamics, and subject consistency in the synthesized videos. For the evaluation, MagicDistillation is assessed using a 16:9 resolution ratio, and the performance metrics of the comparison methods are sourced directly from the official leaderboard available at \textcolor{C6}{\href{https://huggingface.co/spaces/Vchitect/VBench\_Leaderboard}{\textcolor{C6}{https://huggingface.co/spaces/Vchitect/VBench\_Leaderboard}}}.

\section{Hyperparameter Settings}
\label{apd:hyperparameter}

\paragraph{LCM.} The LCM algorithm utilized in our training process largely follows the methodology outlined in FastVideo~\citep{fastvideo}. Specifically, we set the number of timesteps for training to 28, with a batch size of 16 (equivalent to performing gradient accumulation four times). The learning rate is fixed at 1e-6 throughout the training process, and loss matching is conducted at $t=0$ using the Huber loss function. It is important to note that, due to GPU memory constraints, we adopt the same approach as FastVideo by not employing EMA. Instead, we treat the target model and the teacher model as identical during LCM training. For evaluation, we use the checkpoint obtained after completing 20k iterations.

\paragraph{Vanilla DMD/DMD2.} The vanilla DMD/DMD2 converges much faster than the LCM, so the time we spend training the vanilla DMD/DMD2 is relatively short, perhaps only a few hundred to a few thousand iterations. Since there is no work on applying vanilla DMD/DMD2 to HunyuanVideo, we determined the baseline using empirical parameterization and used the naked eye to determine what settings would work best with vanilla DMD/DMD2. To be specific, we employ a learning rate of 3e-6 with a batch size of 4 and do not utilize gradient accumulation (as deepspeed does not support the use of two optimizers, resulting in a gradient accumulation bug). Due to GPU memory constraints, we leverage LoRA to fine-tune $\ve{v}_\theta^\textrm{fake}$. Additionally, the discriminator maintains the single block's architecture from HunyuanVideo. The DMD loss $\mathcal{L}_\textrm{DMD}$ is computed at $t=0$, while the timesteps for the 4-step sampling are set to [1000.0, 937.5, 833.3, 625.0]. The LoRA $\zeta$ and the few-step generator $G_\phi$ are updated at a frequency ratio of 5:1. Importantly, as vanilla DMD/DMD2 often experiences training collapse after a certain period, leading to the production of uncontrollable videos, we select the checkpoint demonstrating the best visual performance prior to the collapse for evaluation. Specifically, for the 1-step and 4-step scenarios, we employ the checkpoint obtained after the 750th iteration and the 300th iteration, respectively, for evaluation.

\paragraph{MagicDistillation.} The majority of the experimental setup for MagicDistillation aligns with that of vanilla DMD/DMD2. Unless specified otherwise in the ablation experiments, we set $\alpha_\textrm{strong}$ and $\alpha_\textrm{weak}$ to 1 and 0.25, respectively, as the default configuration. For the 1-step generator, the pre-trained model is trained with $\alpha_\textrm{weak}=0.25$ and optimized under the regularization loss $\mathcal{L}_\textrm{reg}$ scenario using MagicDistillation. As MagicDistillation significantly mitigates the problem of training collapse, we select checkpoints obtained after the 600th iteration for the 4-step scenario and after the 1500th iteration for the 1-step scenario for evaluation.

\section{Proof of Proposition~\ref{the:w2s_dmd_work}.}
\label{apd:the_1}

The section will demonstrate that \textit{W2S} distribution matching has the same optimization objective as vanilla DMD. We first rewrite Eq.~\ref{eq:diffusion_loss} based on \textit{W2S} distribution matching:\begin{equation}
    \begin{split}
    \mathcal{L}_\textrm{diffusion} & = \mathbb{E}_{t\sim \mathcal{U}[0,1],\epsilon^\prime\sim \mathcal{N}(0,\mathbf{I})}\Big[\alpha_t\Vert \epsilon^\prime - G_\phi(\epsilon) -\ve{v}^\textrm{fake}_\theta(\sigma_t\epsilon^\prime+(1-\sigma_t)G_\phi(\epsilon),t)\Vert_2^2\Big] \\
    &  = \mathbb{E}_{t\sim \mathcal{U}[0,1],\epsilon^\prime\sim \mathcal{N}(0,\mathbf{I})}\Big[\alpha_t\Vert \epsilon^\prime - G_\phi(\epsilon) -\ve{v}^\textrm{pre-train}_\Theta(\sigma_t\epsilon^\prime+(1-\sigma_t)G_\phi(\epsilon),t) - \alpha_\textrm{strong}\zeta(\ve{x}_t,t)\Vert_2^2\Big]\\
    &  \approx \mathbb{E}_{t\sim \mathcal{U}[0,1],\epsilon^\prime\sim \mathcal{N}(0,\mathbf{I})}\Big[\alpha_t\Vert G^*_\phi(\epsilon)  - G_\phi(\epsilon) - \alpha_\textrm{strong}\zeta(\ve{x}_t,t)\Vert_2^2\Big],\\
    \end{split}
    \label{eq:derivation_1}
\end{equation}
where $G_\phi^*(\epsilon) = \epsilon^\prime - \ve{v}^\textrm{pre-train}_\Theta(\sigma_t\epsilon^\prime+(1-\sigma_t)G_\phi(\epsilon),t)$. Thus, minimizing Eq.~\ref{eq:derivation_1} yields
\begin{equation}
    \begin{split}
    & \zeta(\ve{x}_t,t) = \frac{G^*_\phi(\epsilon) - G_\phi(\epsilon)}{\alpha_\textrm{strong}}.\\
    \end{split}
    \label{eq:derivation_2}
\end{equation}
Then, we rewrite the DMD optimization objective (\textit{i.e.}, Eq.~\ref{eq:the_int}) in the main paper:
\begin{equation}
    \begin{split}
        & \mathop{\arg\min}_{\phi}\ \mathbb{E}_{t,(\epsilon,\epsilon^\prime)\sim \mathcal{N}(0,\mathbf{I})}\left[\Vert\epsilon^\prime - G_\phi(\epsilon) - \ve{v}_\Theta^\textrm{real}(\ve{x}_t,t)\Vert_2^2\right]\\
        &\implies \mathop{\arg\min}_{\phi}\ \mathbb{E}_{t,(\epsilon,\epsilon^\prime)\sim \mathcal{N}(0,\mathbf{I})}\left[\Vert\epsilon^\prime - G_\phi(\epsilon) - \ve{v}_\Theta^\textrm{pre-train}(\ve{x}_t,t) - \alpha_\textrm{weak}\zeta(\ve{x}_t,t)\Vert_2^2\right] \\
        &\implies \mathop{\arg\min}_{\phi}\ \mathbb{E}_{t,(\epsilon,\epsilon^\prime)\sim \mathcal{N}(0,\mathbf{I})}\left[\Vert G_\phi^*(\epsilon) - G_\phi(\epsilon) - \alpha_\textrm{weak}\zeta(\ve{x}_t,t)\Vert_2^2\right] \\
        &\implies \mathop{\arg\min}_{\phi}\ \mathbb{E}_{t,(\epsilon,\epsilon^\prime)\sim \mathcal{N}(0,\mathbf{I})}\left[\left\Vert \left(\frac{\alpha_\textrm{strong}-\alpha_\textrm{weak}}{\alpha_\textrm{strong}}\right)[G_\phi^*(\epsilon) - G_\phi(\epsilon)]\right\Vert\right] \\
        &\implies \mathop{\arg\min}_{\phi}\ \mathbb{E}_{t,(\epsilon,\epsilon^\prime)\sim \mathcal{N}(0,\mathbf{I})}\left(\frac{\alpha_\textrm{strong}-\alpha_\textrm{weak}}{\alpha_\textrm{strong}}\right)^2\left[\left\Vert \epsilon^\prime - \ve{v}^\textrm{pre-train}_\Theta(\sigma_t\epsilon^\prime+(1-\sigma_t)G_\phi(\epsilon),t) - G_\phi(\epsilon)\right\Vert_2^2\right] \\
        &\implies \mathop{\arg\min}_{\phi}\ \mathbb{E}_{t,(\epsilon,\epsilon^\prime)\sim \mathcal{N}(0,\mathbf{I})}\left(\frac{\alpha_\textrm{strong}-\alpha_\textrm{weak}}{\alpha_\textrm{strong}}\right)^2\\
        &\left[\left\Vert \ve{v}_\theta^\textrm{fake}(\sigma_t\epsilon^\prime+(1-\sigma_t)G_\phi(\epsilon),t) - \ve{v}^\textrm{pre-train}_\Theta(\sigma_t\epsilon^\prime+(1-\sigma_t)G_\phi(\epsilon),t)\right\Vert_2^2\right]. \\
    \end{split}
    \label{eq:derivation_3}
\end{equation}
Thus, the optimization objective of \textit{W2S} distribution matching is identical to that of the vanilla DMD.

In fact, $\zeta(\ve{x}_t,t)$ usually fails to fit $\frac{G^*_\phi(\epsilon) - G_\phi(\epsilon)}{\alpha_\textrm{strong}}$ completely, so the loss of \textit{W2S} distribution matching eventually translates into a minimization of $(\alpha_\textrm{strong}-\alpha_\textrm{weak})^2\Vert\zeta(\ve{x}_t,t)\Vert_2^2$ as well. In this scenario, the low-rank branch $\zeta(\ve{x}_t,t)$ functions as an intermediate teacher, effectively acting as an overload mechanism that enhances the process of knowledge distillation (\textit{i.e.}, \textit{W2S} distribution matching). Furthermore, while it is often stated that $\alpha_\textrm{weak}$ and $\alpha_\textrm{strong}$ serve merely as weighting factors in Eq.~\ref{eq:derivation_3}, the primary role of $\alpha_\textrm{weak}$ is, in fact, to prevent $G_\phi(\epsilon)$ from exceeding the permissible input range of $\ve{v}_\Theta^\textrm{pre-train}$. This safeguard is crucial, as exceeding this range could lead to instability during training. Notably, this more conservative estimation of the gradient direction, $\nabla \log p_\textrm{fake} - \nabla \log p_\textrm{real}$, results in tangible performance improvements. Through our experiments, we observe that setting $\alpha_\textrm{weak} = 0.25$ achieves the optimal balance, outperforming the vanilla DMD approach.

\section{Why We Choice DMD?}
\label{apd:choice_dmd}

\begin{figure*}[t]
        \centering
        \includegraphics[width=\linewidth,trim={0 0 0cm 0},clip]{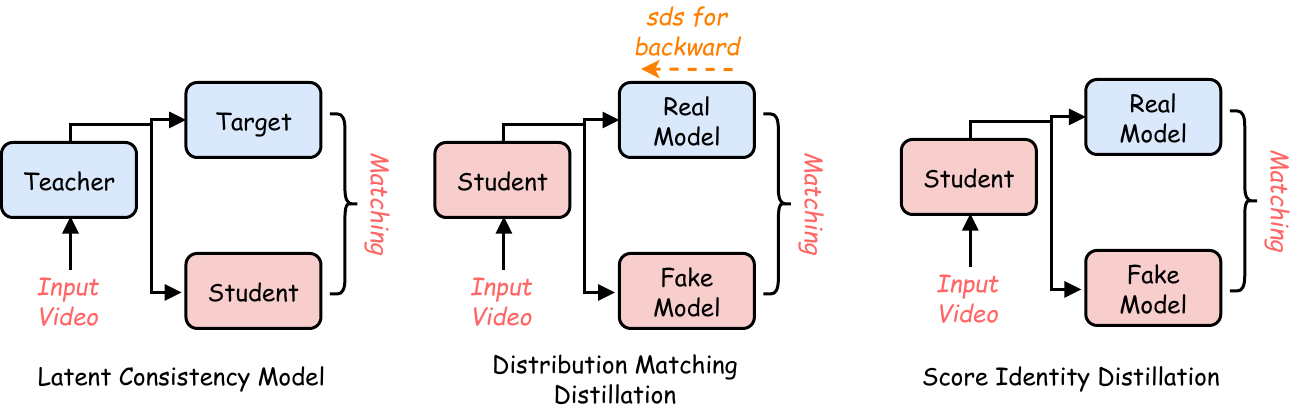} 
        \vspace{-15pt}
        \caption{Comparison of LCM, DMD and SiD training paradigms.}
        \label{fig:sid_dmd_lcm}
        \vspace{-10pt}
\end{figure*}

Most existing methods for step distillation in VDMs are built upon the latent consistency model (LCM) paradigm~\citep{icml23_consistency}. However, as illustrated in Fig.~\ref{fig:sid_dmd_lcm}, the LCM framework necessitates the incorporation of the exponential moving average (EMA) mechanism, resulting in the requirement to load three models' parameters during training. Unfortunately, through our exploratory studies, we observed that even with the use of 16 GPUs equipped with 80GB and the zero redundancy optimizer (ZeRO) stage 3 CPU offload technique~\citep{zero3}, the out-of-memory (OOM) issue persists and cannot be resolved.

Consequently, we turn our attention to an EMA-independent step distillation method known as score identity distillation (SiD)~\citep{sid_lsg} and distribution matching distillation (DMD)~\citep{dmd,dmd2}. Compared to DMD, SiD necessitates calculating the gradients of both the real model and the fake model during the update of the few-step generator, resulting in significantly lower computational efficiency. Consequently, we ultimately chose DMD, which is expected to enable the development of the few-step generator, which theoretically achieves both efficiency and effectiveness. In detail, DMD aims to estimate both the real data distribution and the fake data distribution using two distinct models with non-overlapping parameters. The generator's parameters are then updated using the score distillation sampling (SDS) algorithm~\citep{iclr2023_dreamfusion}, leveraging information from both distributions. It is clear that the training process of vanilla DMD similarly requires three models, which can result in the same OOM issue as encountered with LCM. To address this issue without compromising the generator's final performance, we propose adding the low-rank adaptation (LoRA) on the real model (\textit{i.e.}, estimate real data distribution) to implement the fake model (\textit{i.e.}, estimate fake data distribution), ensuring efficient training and avoiding the OOM issue.

\section{Experimental Result of Magic141-TI2V}
\label{apd:ti2v_generator}

\begin{figure*}[!h]
    \vspace{-5pt}
    \centering
    \includegraphics[width=1.0\linewidth]{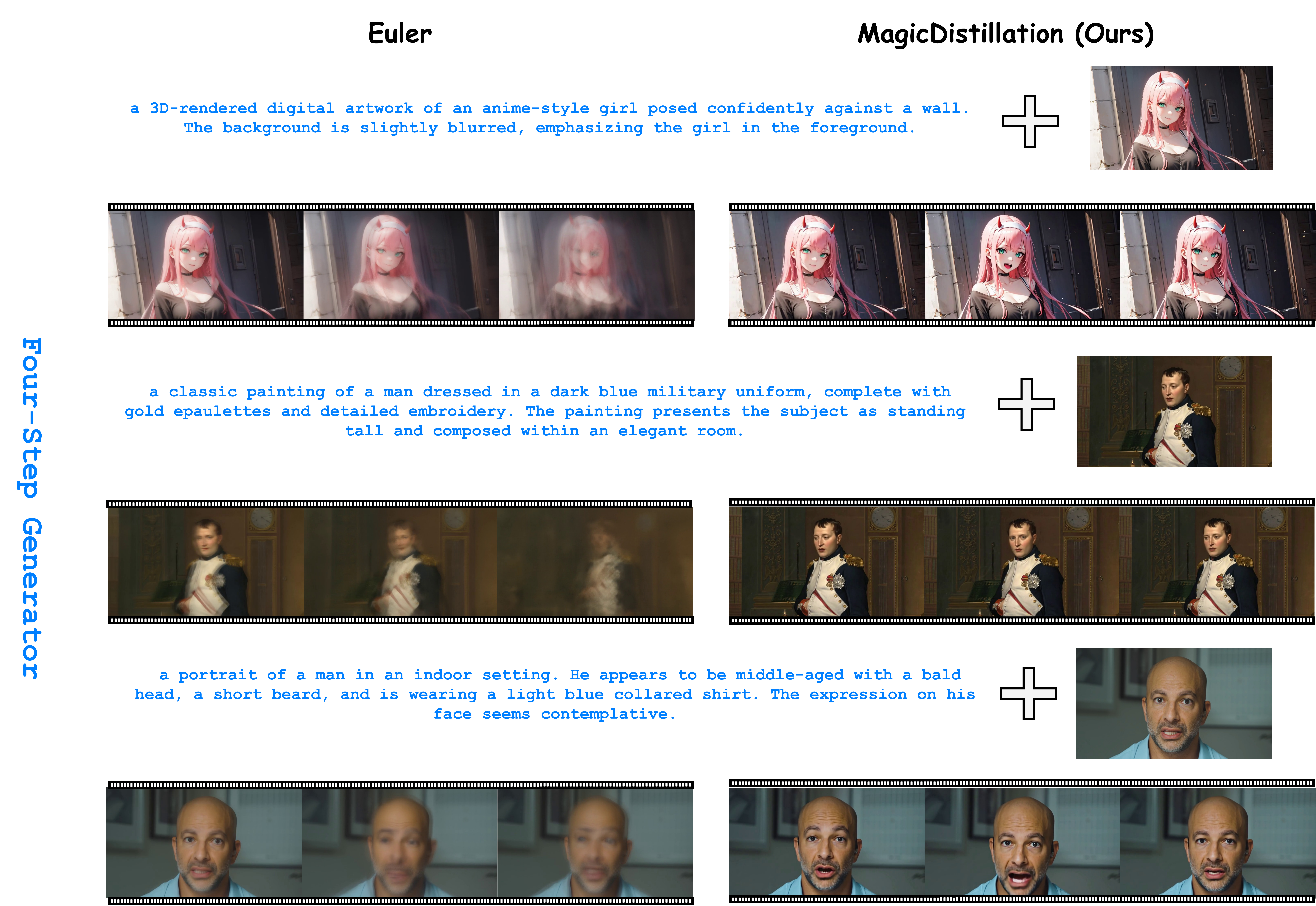}
    \vspace{-15pt}
    \caption{Visualization of MagicDistillation and the standard Euler sampling on the TI2V-HunyuanVideo model. In the 4-step scenario, MagicDistillation visibly outperforms Euler in terms of visual quality and facial motion dynamics.}
    \label{fig:post_visualization_apd}
    \vspace{-10pt}
\end{figure*}

Herein, we present the performance of MagicDistillation on TI2V-HunyuanVideo. Following the identical experimental setup employed for MagicDistillation in I2V-HunyuanVideo, we enhanced computational resources to 8 nodes with 40 NVIDIA H100 GPUs, aiming to achieve superior performance. As demonstrated in Fig.~\ref{fig:post_visualization_apd} and Table~\ref{tab:comparison_apd} containing visualizations and quantitative evaluations, the distilled 4-step generator through MagicDistillation exhibits enhanced synthesis fidelity and improved motion consistency in TI2V portrait video generation tasks. These empirical results comprehensively validate the technical advantages of MagicDistillation, particularly highlighting its capability to preserve temporal coherence while maintaining high visual quality in compressed-step video synthesis scenarios.

\begin{table*}[!t]
\centering
\vskip -0.01in
\small
\caption{\small {TI2V quantitative results comparison using Magic141 (trained on both portrait and general video datasets) on our customized VBench. All videos are 540$\times$960$\times$129 and are saved at 24fps. Each sample follow VBench~\citep{vbench} to synthesize 5 videos to avoid errors.}}
\scalebox{0.63}{
\begin{tabular}{lccccccccc}
\Xhline{3\arrayrulewidth}\\[-2ex]
{\textbf{Method}} & {\textbf{\makecell{i2v\\subject} (↑)}} & {\textbf{\makecell{subject\\consistency} (↑)}} & {\textbf{\makecell{motion\\smoothness} (↑)}} & {\textbf{\makecell{dynamic\\degree} (↑)}} & {\textbf{\makecell{aesthetic\\quality} (↑)}} & {\textbf{\makecell{imaging\\quality} (↑)}} & {\textbf{\makecell{temporal\\flickering} (↑)}} &  {\textbf{average (↑)}}\\
\Xhline{3\arrayrulewidth}
      \\[-2ex]
        \multicolumn{9}{l}{\large\textit{\textbf{28-step}}}\\\Xhline{3\arrayrulewidth}
{Euler (baseline)} & 0.9274 & 0.9397 & 0.9953 & 0.2448 & 0.5687 & 0.6671 & 0.9935 &  0.7623 \\
\\[-2ex]
\multicolumn{9}{l}{\large\textit{\textbf{4-step}}}\\\Xhline{3\arrayrulewidth}
{Euler (baseline)} & 0.9803 & 0.8593 & \CC 0.9965 & 0.0034 & 0.4440 & 0.3693 & 0.9972 &  0.6642 \\
{MagicDistillation w/ reg. loss (ours)} & \CC 0.9807 & \CC 0.9777 & 0.9951 & \CC 0.0620 & \CC 0.5979 & \CC 0.7110 & \CC 0.9974 & \CC 0.7602 \\
\Xhline{3\arrayrulewidth}\\[-2ex]
\end{tabular}
}
\label{tab:comparison_apd}
\vskip -0.07in
\end{table*}

\section{Pseudo Code of MagicDistillation}
\label{apd:pseudo_code}

In order to give the researchers a better understanding of MagicDistillation, we have placed the pseudo-code of this method in Algorithm~\ref{alg:distillation}.

\begin{algorithm}
    \scriptsize
    \caption{\label{alg:distillation} \textbf{MagicDistillation} Training Procedure}
    \KwIn{Pretrained Hunyuan VDM $\ve{v}_\Theta^\textrm{pre-train}$,
    the ground truth video dataset $\mathcal{D}=\{\ve{x}^\textrm{gt}_i\}$, the low-rank branch $\zeta$, the number of denoising steps $N$, the denoising timestep list $[t_1,\cdots,t_N]$.}
    \KwOut{Trained generator $G_\phi$.}
    \tcp{Initialize the few-step generator and the real data distribution estimator $\ve{v}_\Theta^\textrm{real}$, the fake data distribution estimator $\ve{v}_\theta^\textrm{fake}$}
    $G_\phi \leftarrow \text{copyWeights}(\ve{v}_\Theta^\textrm{pre-train}),$
    $ \ve{v}^\textrm{real}_\Theta \leftarrow \alpha_\textrm{weak} \zeta + \ve{v}_\Theta^\textrm{pre-train},$
    $ \ve{v}^\textrm{fake}_\theta \leftarrow \alpha_\textrm{strong} \zeta + \ve{v}_\Theta^\textrm{pre-train} $\\
    \While{train}{
        \tcp{Generate videos}
        Sample batch $\epsilon \sim \mathcal{N}(0, \mathbf{I})^B$, the ground truth video $\ve{x}^\textrm{gt} \sim \mathcal{D}$ and $t \sim [t_1,\cdots,t_N]$        \text{~}
        \tcp{Prepare for MagicDistillation}
        Get the noisy video $\ve{x}_t = (t/1000)\epsilon + ((1-t)/1000)\ve{x}^\textrm{gt}$
        \text{~}
        \tcp{Prepare for MagicDistillation}
        Get the synthesized noisy video $\ve{\tilde{x}}_t = (t/1000)\epsilon + ((1-t)/1000)[\ve{x}_t - tG_\phi(\ve{x}_t, t)]$ 
        \text{~}
        \tcp{Prepare for MagicDistillation}
        $\mathcal{L}_\text{DMD} \leftarrow \text{distributionMatchingLoss}(\ve{v}^\textrm{real}_\Theta, \ve{v}^\textrm{fake}_\theta, \ve{\tilde{x}}_t)$ \tcp{Eq~\ref{eq:dmd_loss}}
        $\mathcal{L}_\text{reg} \leftarrow \text{distributionMatchingLoss}(\ve{x}^\textrm{gt}, \ve{v}^\textrm{fake}_\theta, \ve{\tilde{x}}_t)$ \tcp{Eq~\ref{eq:reg_loss}}
        $\mathcal{L}_\text{gen} \leftarrow \text{generatorLoss}(D_\xi,G_\phi,\epsilon, \ve{x}^\textrm{gt})$ \tcp{Eq~\ref{eq:adv_loss}}
        $\mathcal{L}_{G_\phi} \leftarrow \mathcal{L}_\text{DMD} + \mathcal{L}_\text{reg} + \mathcal{L}_\text{dis} $\text{~}\tcp{Get the overall loss}
        $G_\phi \leftarrow \text{update}(G_\phi, \mathcal{L}_{G_\phi})$
        \text{~}
        \tcp{Update the few-step generator}
        Sample time step $t\sim\mathcal{U}[0,1]$ and the Gaussian noise $\epsilon \sim \mathcal{N}(0, \mathbf{I})^B$

        $\ve{x}^\textrm{fake}_0 \leftarrow \ve{x}_t - tG_\phi(\ve{x}_t, t)$
        
        $\ve{x}^\textrm{fake}_t \leftarrow (t/1000)\epsilon + ((1-t)/1000)\ve{x}^\textrm{fake}_0$

        $ \mathcal{L}_{\text{diffusion}} \leftarrow \text{diffusionLoss}(\ve{v}^\text{fake}_\theta(\ve{x}^\textrm{fake}_t, t), \text{stopgrad}(\ve{x}^\textrm{fake}_0))$ \tcp{Eq~\ref{eq:diffusion_loss}}
        $\mathcal{L}_\text{dis} \leftarrow \text{discriminatorLoss}(D_\xi,G_\phi,\epsilon)$ \tcp{Eq~\ref{eq:adv_loss}}
        
        $\zeta \leftarrow \text{update}(\zeta, \mathcal{L}_{\text{diffusion}}+\mathcal{L}_\text{dis})$\text{~}
        \tcp{Update the low-rank branch}
    }
\end{algorithm}

\section{Training Collapse of Vanilla DMD}
\label{apd:training_collapse_dmd}

\begin{figure*}
    \centering
    \includegraphics[width=1.0\linewidth]{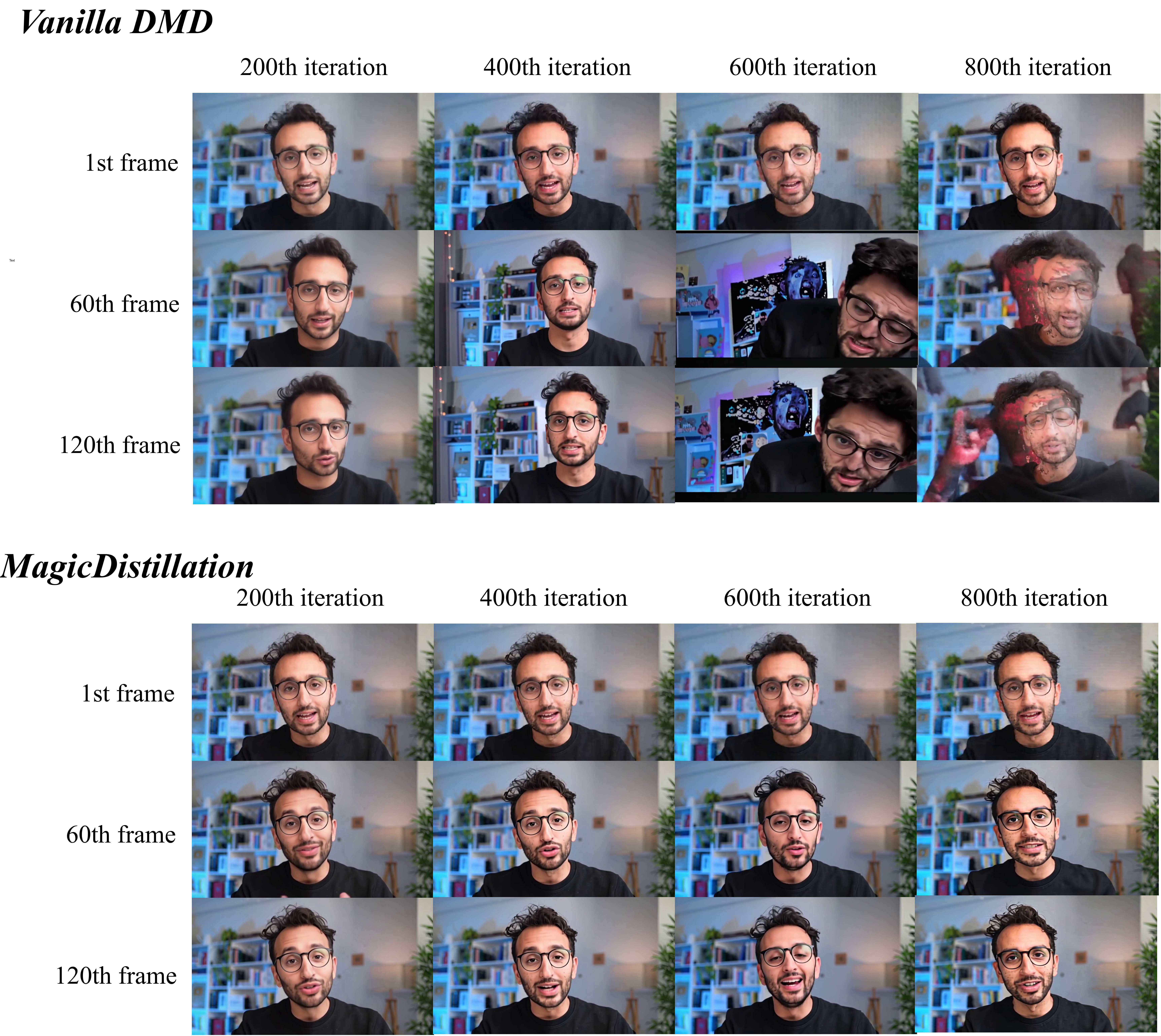}
    \caption{Comparison between vanilla DMD and MagicDistillation in terms of training stability. It is worth noting that by the 400th iteration, vanilla DMD exhibits a noticeable change in the background, which serves as an early warning sign of impending training collapse. In contrast, MagicDistillation (w/ reg. loss) demonstrates enhanced stability, effectively mitigating such issues throughout the training process.}
    \label{fig:training_collapse}
\end{figure*}
Here, we employ visualization to demonstrate the training collapse issue encountered with vanilla DMD. It is important to highlight that the number of iterations required to distill HunyuanVideo from 28 steps to 4 steps is remarkably small, demanding only a few hundred to a few thousand iterations with a batch size of 4. However, vanilla DMD suffers from training collapse after approximately 300 iterations, leading to the synthesis of uncontrollable videos. In Fig.~\ref{fig:training_collapse}, we present the videos synthesized by vanilla DMD in the 4-step scenario across different numbers of iterations, where this collapse phenomenon is clearly observable. In contrast, MagicDistillation effectively addresses this issue by aligning the real data distribution with the fake data distribution, while also leveraging ground truth supervision to entirely avoid this problem. As shown in {In Fig.~\ref{fig:training_collapse}, MagicDistillation consistently produces high-quality portrait videos throughout the entire training process.

\section{Our Customized VBench}

\begin{figure*}
    \centering
    \includegraphics[width=1.0\linewidth]{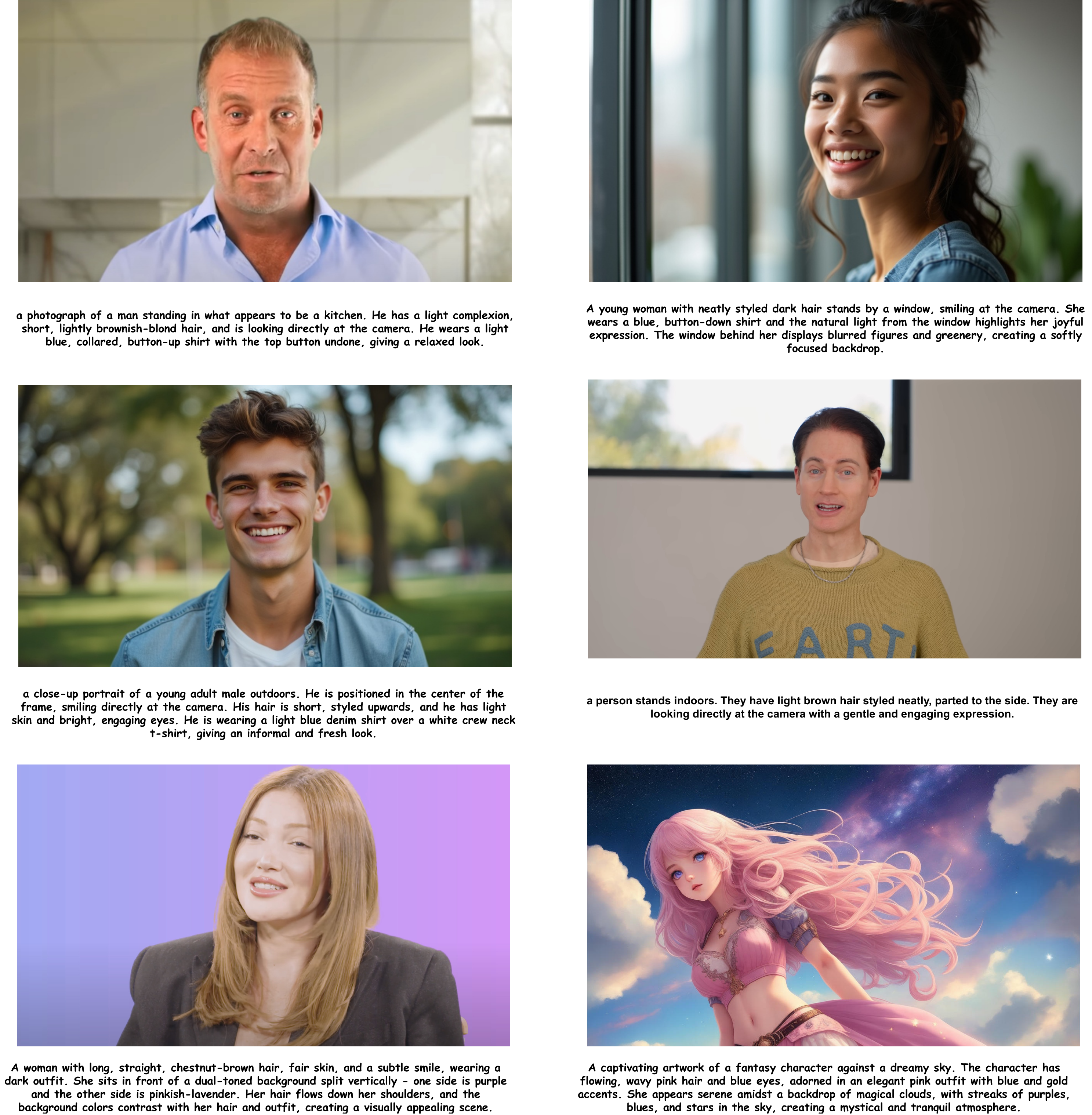}
    \caption{Our tailored VBench is composed of widescreen reference images paired with their corresponding text prompts. These reference images encompass both animated and real characters, amounting to a total of 58 samples. To ensure robust evaluation and minimize errors stemming from randomness, we synthesize five videos for each sample during the evaluation of our customized VBench.}
    \label{fig:vbench_example}
\end{figure*}
To justify the effectiveness of our VBench designed for the portrait video synthesis task, we present several illustrative use cases within our customized VBench. As depicted in Fig.~\ref{fig:vbench_example}, the reference images in these examples span both real-life and animated characters, with the accompanying text prompts providing clear and descriptive information about the reference images.

\section{Additional Visualization}
\label{apd:add_visualization}

\begin{figure*}[!t]
    \vspace{-5pt}
    \centering
    \includegraphics[width=1.0\linewidth]{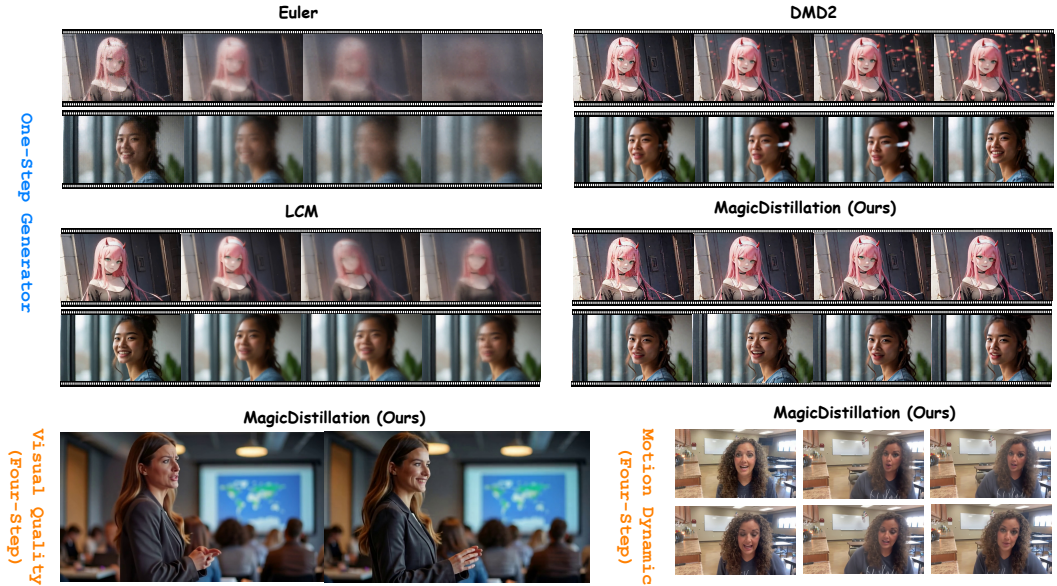}
    \vspace{-15pt}
    \caption{Visualization of various accelerated sampling methods under the 1-step scenario. The video quality and dynamic degree synthesized by MagicDistillation are visually superior to those produced by Euler, LCM, and vanilla DMD2.}
    \label{fig:post_visualization_2}
    \vspace{-10pt}
\end{figure*}

\begin{figure*}
    \centering
    \includegraphics[width=1.0\linewidth]{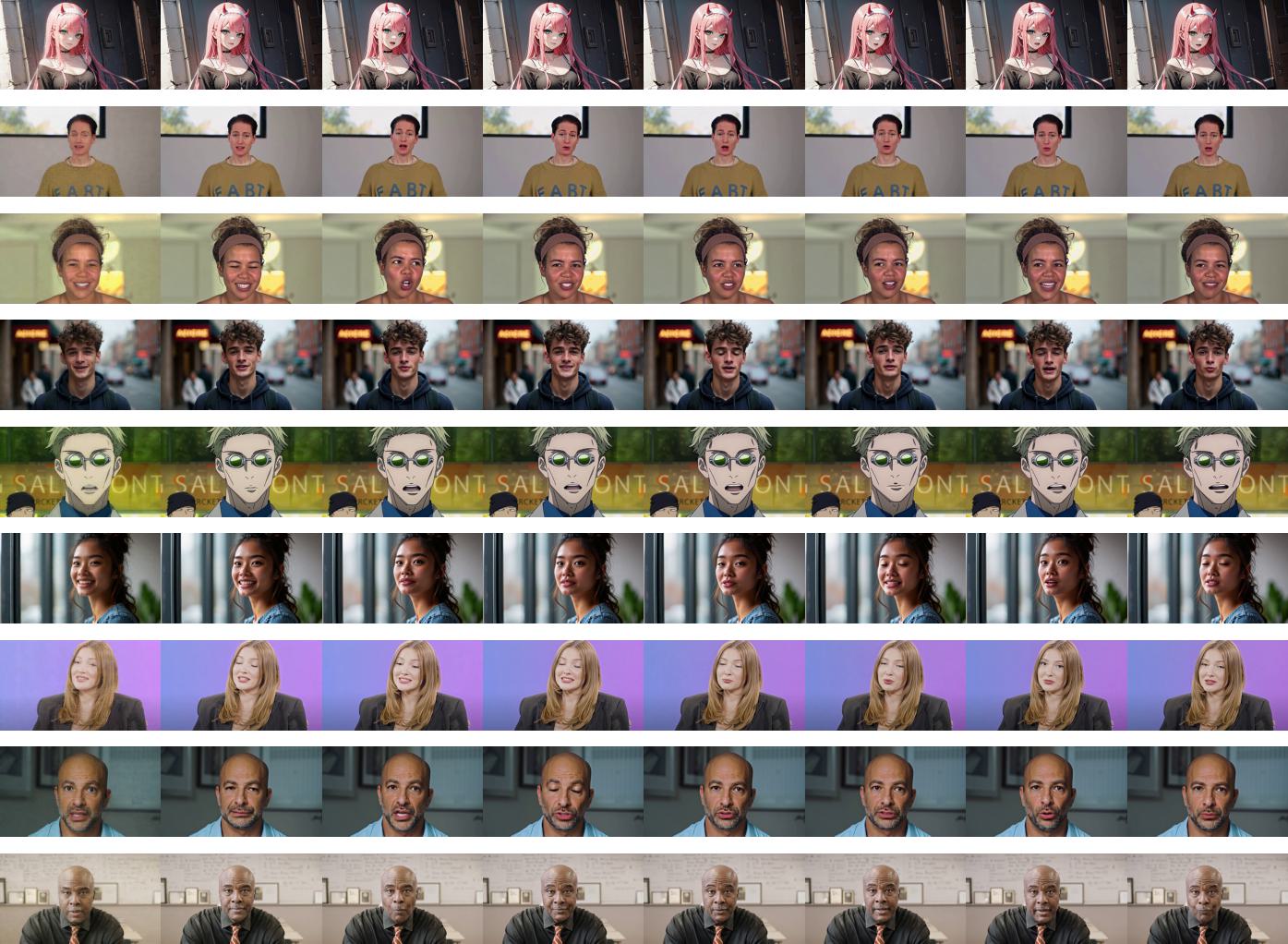}
    \caption{Visualization of MagicDistillation at the 4-step scenario. The leftmost frame of the video (\textit{i.e.}, the first frame) is the reference image. It can be seen that all videos have high quality and the 4-step generator does not synthesize uncontrollable videos.}
    \label{fig:vis_w2s_dmd}
\end{figure*}

\begin{figure*}
    \centering
    \includegraphics[width=1.0\linewidth]{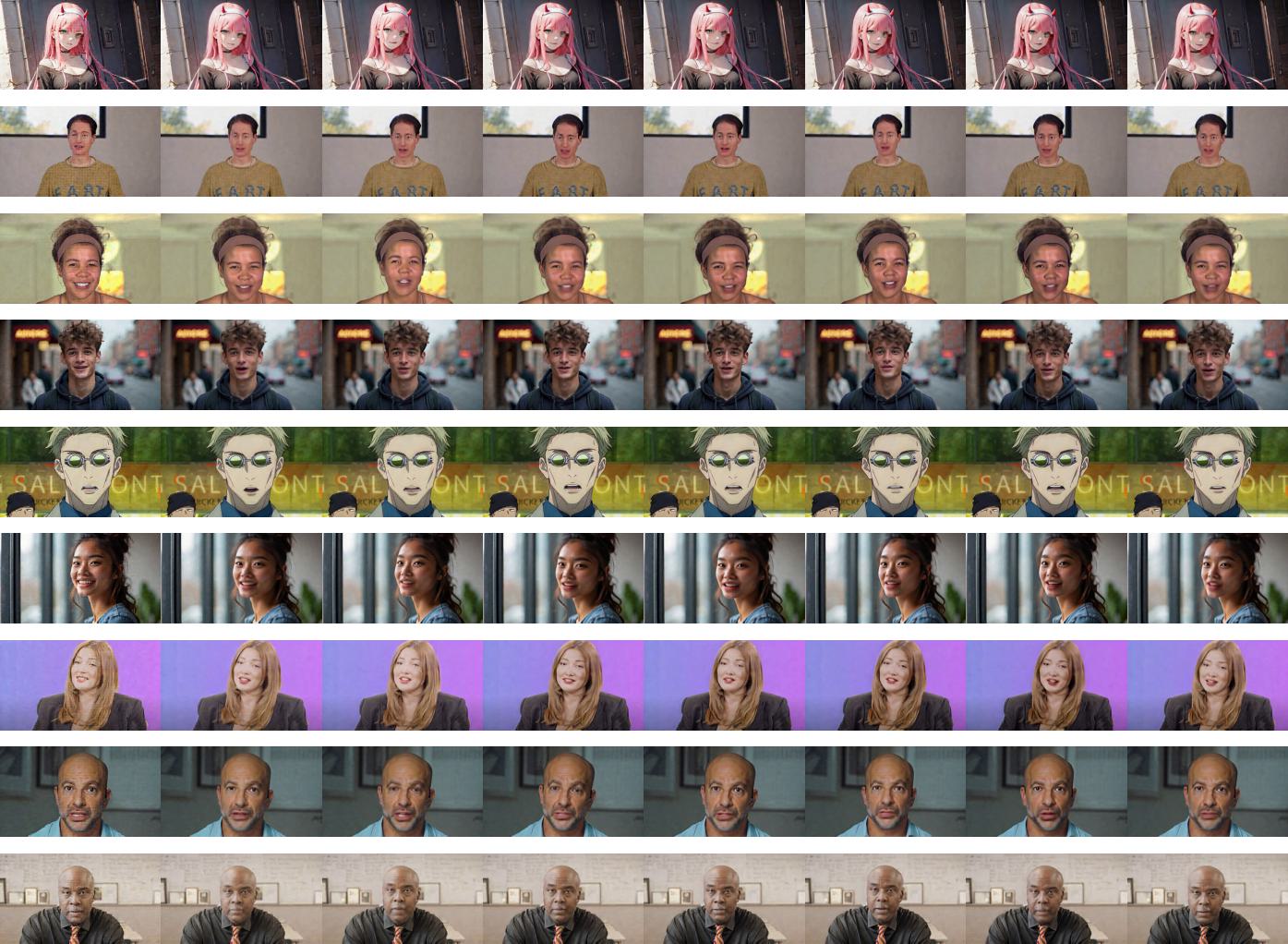}
    \caption{Visualization of MagicDistillation at the 1-step scenario. The leftmost frame of the video (\textit{i.e.}, the first frame) is the reference image. It can be seen that all videos have relatively high quality and the 1-step generator does not synthesize uncontrollable videos.}
    \label{fig:vis_w2s_dmd_one_step}
\end{figure*}

Here, we first present the video frames under the 1-step scenario and further showcase additional video frames from the synthesized videos of MagicDistillation. For the 1-step scenario, as shown in Fig. ~\ref{fig:post_visualization_2}, the visual quality of MagicDistillation remains significantly better than that of the comparison methods, despite a slight decrease compared to the 4-step scenario. For more examples, the complete videos can be accessed on {\textcolor{C6}{\href{https://magicdistillation.github.io/MagicDistillation/}{\textcolor{C6}{https://magicdistillation.github.io/MagicDistillation/}}}}. Specifically, we provide video frames from two categories of synthesized videos: the 1-step and 4-step scenarios. The visualization results for the 1-step scenario are presented in Fig.~\ref{fig:vis_w2s_dmd_one_step}, while those for the 4-step scenario are displayed in Fig.~\ref{fig:vis_w2s_dmd}.

\end{document}